\documentclass{article} 
\usepackage{iclr2027_conference,times}

\usepackage{amsmath,amsfonts,bm}

\def\eqref#1{equation~\ref{#1}}

\def\1{\bm{1}}

\DeclareMathAlphabet{\mathsfit}{\encodingdefault}{\sfdefault}{m}{sl}
\SetMathAlphabet{\mathsfit}{bold}{\encodingdefault}{\sfdefault}{bx}{n}

\usepackage{hyperref}
\usepackage{url}
\usepackage{booktabs} 
\usepackage{graphicx}
\usepackage{float}
\usepackage{wrapfig}
\usepackage[table]{xcolor}
\usepackage[most]{tcolorbox}

\newtcblisting{promptbox}[1]{
    enhanced,
    breakable,
    listing only,
    listing engine=listings,
    title={#1},
    colback=black!2,
    colframe=black!35,
    coltitle=black,
    fonttitle=\bfseries\small,
    boxrule=0.5pt,
    arc=1mm,
    left=1.5mm,
    right=1.5mm,
    top=1mm,
    bottom=1mm,
    before skip=5pt,
    after skip=5pt,
    listing options={
        basicstyle=\ttfamily\scriptsize,
        breaklines=true,
        columns=fullflexible,
        keepspaces=true,
        showstringspaces=false
    }
}

\definecolor{BestInModule}{RGB}{235,245,255}
\definecolor{BestOverall}{RGB}{235,255,235}
\definecolor{LightOrange}{RGB}{255,248,235}

\title{WorldLine: Action-Driven Visual Simulation for Robotic Manipulation}

\author{%
\parbox[t]{\dimexpr\textwidth-2\tabcolsep\relax}{%
\raggedright
\small
Shenghe Zheng\textsuperscript{1}\hspace{0.45em}
Wenbo Li\textsuperscript{2}\thanks{Project Lead.}\hspace{0.45em}
Jiyao Zhang\textsuperscript{3}\hspace{0.45em}
Bin Xia\textsuperscript{4}\hspace{0.45em}
\textbf{Haoyang Huang}\textsuperscript{2}\hspace{0.45em}
\textbf{Nan Duan}\textsuperscript{2}\hspace{0.45em}
\textbf{Jiaya Jia}\textsuperscript{1}\thanks{Corresponding author.}\endgraf
\vspace{0.45em}
\normalfont\small
\mbox{\textsuperscript{1}The Hong Kong University of Science and Technology}\hspace{2em}
\mbox{\textsuperscript{2}Joy Future Academy}\hspace{2em}
\mbox{\textsuperscript{3}Peking University}\hspace{2em}
\mbox{\textsuperscript{4}The Chinese University of Hong Kong}
}%
}

\iclrfinalcopy 
\begin{document}

\maketitle

\begin{figure}[H]
    \centering
    \vspace{-0.5cm}
    \includegraphics[width=\textwidth]{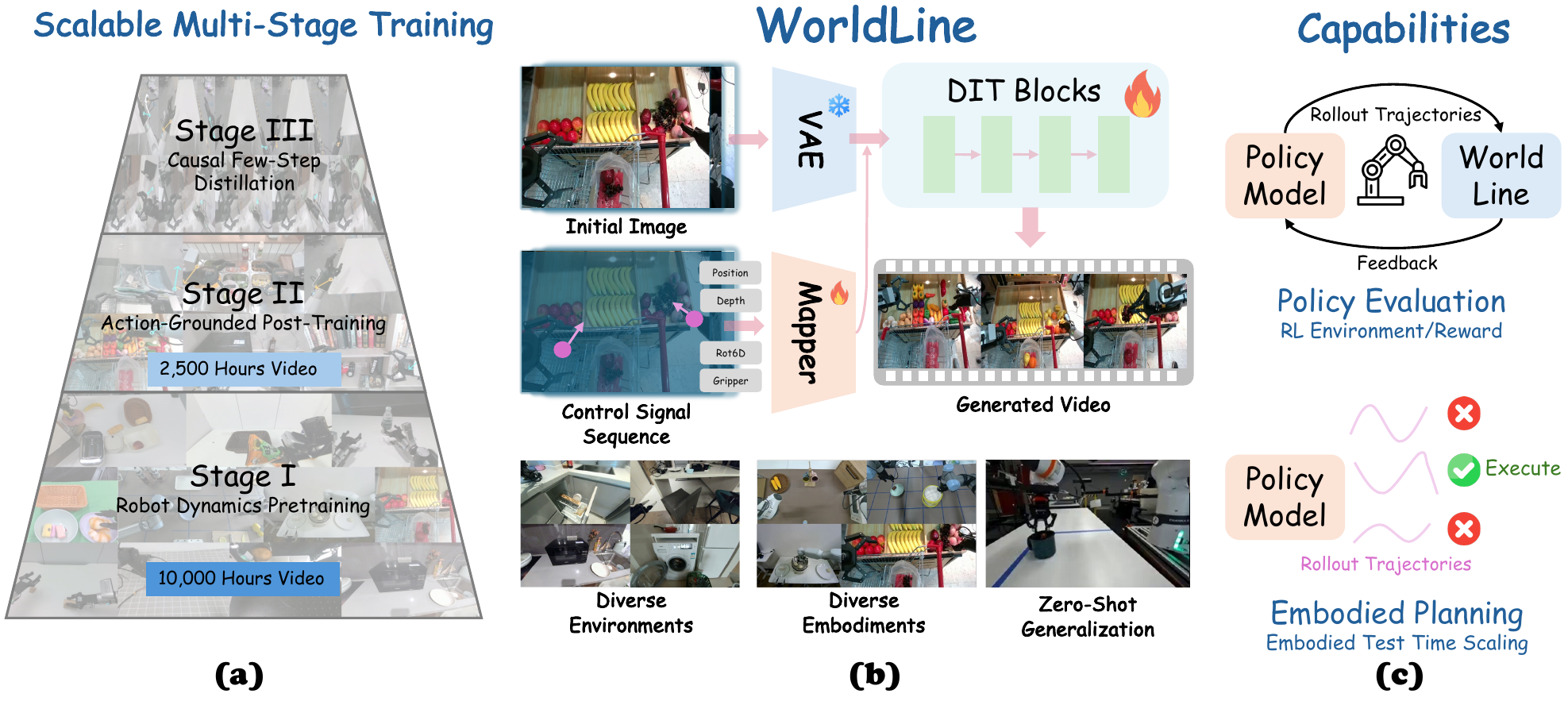}
    \vspace{-0.5cm}
    \caption{Overview of \textbf{WorldLine}. (a) A scalable three-stage training
    pipeline: \textbf{Stage I} learns robot dynamics from large-scale action-free
    videos; \textbf{Stage II} grounds heterogeneous robot controls through
    action-conditioned post-training; and \textbf{Stage III} distills the model
    for causal few-step rollout.
    (b) WorldLine maps an initial observation and an image-space control sequence
    to future videos across diverse environments and robot embodiments,
    including zero-shot settings. (c) WorldLine support policy
    evaluation and planning by predicting action outcomes before
    real-world execution.}
    \label{fig:teaser}
\end{figure}

\begin{abstract}
Real-world robot learning is constrained by the cost of collecting experience
and evaluating candidate behaviors. Video generation models offer a scalable
foundation for visual simulators that predict action outcomes before physical
execution. Yet they often favor visual plausibility over accurate action
following and coherent robot--object dynamics, while action-conditioned
simulators depend on scarce, embodiment-specific data that are difficult to
share across incompatible control spaces. We introduce \textbf{WorldLine}, an
action-driven visual simulator that decouples transferable dynamics learning
from heterogeneous action grounding. WorldLine learns manipulation dynamics
from more than 10,000 hours of action-free robot videos and grounds them using
over 2,000 hours of action trajectories across more than ten embodiments. An
image-space action representation provides a shared control interface across
embodiments, while multi-view and failure-enriched training with relational
regularization improves interaction-sensitive prediction. Robot-focused few-step
distillation enables efficient causal rollout while preserving action-critical
motion. Across held-out and out-of-domain settings, WorldLine maintains strong
visual quality and robot-motion agreement; on failed trajectories, it improves
robot-mask IoU by 0.1626 over the strongest baseline. It predicts trajectory
success with 74\% mean accuracy across RoboTwin and AgiBot, one percentage point
above the strongest baseline. Without RoboTwin training or adaptation, its rollouts
improve task success by up to 21.4 percentage points
over direct policy execution. Together, these capabilities make WorldLine a
scalable and efficient visual simulator for policy evaluation and embodied planning.
More results are available at \href{https://zhengsh123.github.io/WorldLine/}{project page}.
\end{abstract}

\section{Introduction}

Robot learning demands large-scale interaction data~\cite{RoboNet,pi0,dreamzero},
yet real-world collection is costly. Conventional
simulators~\cite{Mujoco,isaac} reduce this burden but require handcrafted assets
and dynamics, limiting scalability across environments and embodiments.
Video generation models~\cite{sora,wan} offer a data-driven visual prior, but
their pretraining objectives primarily emphasize visual realism and temporal
consistency without explicitly enforcing faithful responses to robot controls.
Action-conditioned visual simulation therefore requires \emph{action fidelity}:
generated trajectories must accurately follow commanded direction, timing, and
gripper state while preserving coherent robot--object dynamics over
time~\cite{ctrl-world,cosmos3,iteractive}.

Learning a visual simulator that is both action-faithful and physically
coherent at scale is difficult. Such a model requires large and diverse robot
experience, yet action supervision is scarce and embodiment-specific. Pooling
trajectories across robots is further complicated by native control signals
that differ in dimensionality, kinematics, coordinate systems, and interfaces,
preventing direct sharing~\cite{maskedvisual,imac,actionimage}.

We introduce \textbf{WorldLine}, an action-driven visual simulator built on the
insight that robot--object dynamics and precise action grounding can be learned
separately, allowing dynamics learning to scale beyond scarce action-labeled
trajectories. WorldLine first adapts a pretrained video model to more than
10,000 hours of action-free robot videos to learn a transferable prior over
robot motion and object interaction. It then uses over 2,000 hours of action
trajectories spanning more than ten robot embodiments to learn how heterogeneous
controls drive future motion. Rather than forcing incompatible controls into a
common vector space, WorldLine projects their spatial effects, including
end-effector pose and gripper state, into each camera view, providing a shared
image-space interface that preserves action geometry across embodiments. This
action grounding enables predicted motion to faithfully follow commanded
actions.

Accurate control following alone is not enough; predictions should also
distinguish successful and unsuccessful interaction outcomes. WorldLine uses
synchronized multi-view videos to provide complementary observations of robot
and object motion, while failure trajectories expose the model to unsuccessful
outcomes rather than biasing generation toward successful completion.
Relational regularization further promotes temporal and cross-view consistency. Finally, robot-focused
few-step distillation turns the model into an efficient causal simulator without
sacrificing action-critical motion, enabling policy evaluation and embodied
planning.

We evaluate WorldLine from three complementary perspectives:
action-conditioned video generation, policy evaluation, and embodied planning.
For action-conditioned generation, WorldLine maintains strong visual quality
and robot-motion agreement across held-out and out-of-domain settings. On
failed trajectories, it improves robot-mask IoU by 0.1626 over the strongest
baseline. For policy evaluation, WorldLine rollouts, evaluated by Qwen3VL-8B
using extracted visual evidence and deterministic task-specific rules, achieve
74\% mean trajectory-success classification accuracy across
RoboTwin~\cite{robotwin} and AgiBot~\cite{Agibotworld}, compared with 73\% for
the strongest baselines. For RoboTwin planning, WorldLine is evaluated without
RoboTwin training, adaptation, or checkpoint selection. Selecting among
policy-generated candidates using WorldLine rollouts and a Qwen3VL-8B selector
improves task success at $N=32$ by 19.1 and 21.4 percentage points for
$\pi_{0.5}$~\cite{pi0.5} and LingBot-VLA~\cite{lingbot-vla}, respectively,
over direct policy execution. Together, these results demonstrate WorldLine's scalability and
practical value for robot evaluation and planning.

Our contributions are threefold:
\begin{list}{\textbullet}{%
    \setlength{\leftmargin}{1.5em}%
    \setlength{\labelwidth}{0.5em}%
    \setlength{\labelsep}{0.5em}%
}
    \item \textbf{Scalable cross-embodiment learning.} We introduce a
    decoupled framework that learns transferable robot--object dynamics from
    large-scale action-free videos and grounds heterogeneous robot controls
    through a geometry-preserving image-space interface.

    \item \textbf{Interaction-oriented modeling.} We combine synchronized
    multi-view observations, failure-enriched trajectories, and relational
    regularization to improve action-conditioned outcome modeling across
    viewpoints, environments, and embodiments. Few-step distillation further
    enables efficient causal rollout while preserving action-critical robot motion.

    \item \textbf{Broad empirical validation.} We evaluate WorldLine on
    action-conditioned video generation, policy evaluation, and best-of-$N$
    trajectory selection, demonstrating generalization to unseen environments
    and embodiments and consistent improvements over direct policy execution.
\end{list}

\section{Related Work}

\noindent\textbf{Video generation for robot learning.}
Video generation models have supported robot learning through visual prediction,
synthetic data generation, and simulation~\cite{Deepvisualforesight,RoboNet,InteractiveReal-World}.
Existing work uses generated videos to facilitate policy learning and augment
training data~\cite{iteractive,irasim}, or to predict the outcomes of candidate
actions for planning and policy evaluation. The latter requires generated
trajectories to respond accurately to robot controls and preserve coherent
robot--object interactions, rather than merely appear visually
plausible~\cite{ctrl-world,quevedo2026worldgym}. WorldLine targets this predictive setting, with an
emphasis on scalable dynamics learning and action-faithful simulation.

\noindent\textbf{Action conditioning and scalable visual simulation.}
Existing visual simulators represent robot controls using native action
vectors~\cite{cosmos3}, learned latent actions~\cite{lapa,genie}, or spatial
conditions~\cite{imac,maskedvisual}. Native actions provide direct control but
are tied to embodiment-specific kinematics and coordinate systems. Latent
actions can capture motion from unlabeled videos but require additional
grounding before they can represent executable controls. Spatial conditions
provide a geometrically aligned interface, but action representation alone does
not resolve the data-scaling problem. Existing simulators still rely primarily
on limited action-labeled trajectories whose control spaces and interaction
distributions vary across embodiments~\cite{x-embodiment,zimablue}. Most methods
therefore learn visual dynamics and action semantics jointly from the same
trajectories, coupling simulation coverage to available action supervision.
WorldLine instead decouples these objectives by learning transferable
robot--object dynamics from large-scale action-free videos and subsequently
grounding heterogeneous controls through a shared image-space representation.
This design allows action-free videos and embodiment-specific action
trajectories to provide complementary supervision.

\section{Data Pipeline}
\label{sec:data_pipeline}
\begin{table}[t]
    \centering
    \caption{Training-data comparison across action-conditioned robot video
    models in scale, diversity, and action supervision. Embod. denotes the
    number of robot embodiments, Failure denotes the duration of failed
    trajectories, and NR denotes unreported information.}
    \vspace{0.2cm}
    \label{tab:training_data_comparison}
    \scriptsize
    \begin{tabular*}{\linewidth}{@{\extracolsep{\fill}}lcccccl@{}}
        \toprule
        & \multicolumn{2}{c}{\textbf{Scale}}
        & \multicolumn{2}{c}{\textbf{Diversity}}
        & \multicolumn{2}{c}{\textbf{Supervision}} \\
        \cmidrule(lr){2-3}
        \cmidrule(lr){4-5}
        \cmidrule(lr){6-7}
        \textbf{Method} & \textbf{Video-only} & \textbf{Action data} &
        \textbf{Embod.} & \textbf{Tasks} & \textbf{Action repr.} &
        \textbf{Failure} \\
        \midrule
        CTRL-World~\cite{ctrl-world} & N/A & 450 h & 1 & 86 &
        Native Action & 60 h \\
        GE-Sim-V2~\cite{gesim} & N/A & 2,500 h & 1 & NR &
        Image-space map & NR \\
        Masked Visual Action~\cite{maskedvisual} & N/A & 450 h & 1 &
        86 & Masked Action & 60 h \\
        \textbf{Worldline(ours)} & \textbf{10,000 h} &
        \textbf{2,500 h} & \textbf{$>$10} & \textbf{$>$3{,}500} &
        Image-space map & \textbf{200 h} \\
        \bottomrule
    \end{tabular*}
\end{table}

\noindent\textbf{Data composition.}
WorldLine separates dynamics learning and action grounding across
complementary data sources: action-free videos provide interaction
coverage, while geometrically verified action trajectories provide precise
control supervision. Figure~\ref{fig:method} summarizes the data
pipeline, while Table~\ref{tab:training_data_comparison} compares its scale,
diversity, and supervision with action-conditioned video models.

\noindent\textbf{Action-free dynamics corpus.}
We curate more than 10,000 hours of robot videos from six collections,
including AgiBotWorld~\cite{Agibotworld},
RoboCOIN~\cite{robocoin}, the RoboMIND
series~\cite{robomind,robomind2.0}, and Galaxea~\cite{galaxea}. The corpus spans
more than ten robot embodiments and over 3,500 manipulation tasks. We retain one
primary task-facing view per trajectory because camera viewpoints and the number
of available views vary substantially across datasets and robot embodiments.
Stage I trains a text-and-image-to-video (TI2V) model conditioned on the initial
frame and task text. Although some source datasets provide robot states or action
annotations, these signals are not used as model inputs or conditioning in Stage
I; thus, \emph{action-free} refers to the training setup rather than the absence of
such annotations in the source data. We filter out static, reset, corrupted, and
otherwise uninformative clips. Dataset composition and video-sampling details
are provided in Appendix~\ref{app:stage_wise_data} and
Appendix~\ref{app:video_sampling}.

\noindent\textbf{Action-grounding corpus.}
For Stage II, we aggregate over 2,000 hours of action-labeled trajectories
across more than ten robot embodiments. Each trajectory provides temporally
aligned controls and calibrated camera geometry. We use robot kinematics to
project end-effector motion into each camera view and discard samples with
invalid calibration, temporal misalignment, or inconsistent projections. The
verified trajectories are converted into camera-aligned action maps encoding
projected position, depth, Rot6D orientation, and gripper state. We additionally
construct a synchronized multi-view subset containing a head camera and up to
two wrist cameras, and include approximately 200 hours of failure trajectories
to broaden viewpoint and outcome coverage. Action-map construction and
multi-view processing are described in Appendix~\ref{app:action_geometry};
failure-data construction and evaluation separation are detailed in
Appendix~\ref{app:failure_data_evaluation}.

\noindent\textbf{Evaluation separation.}
We reserve a scene-disjoint subset of AgiBotWorld for in-domain evaluation and
exclude it from model training. DROID is excluded from
training and used only to evaluate out-of-distribution generalization across
unseen environments, tasks, and robot embodiments.

\section{Method}

\subsection{Overview}

\begin{figure}[t]
    \centering
    \includegraphics[width=\textwidth]{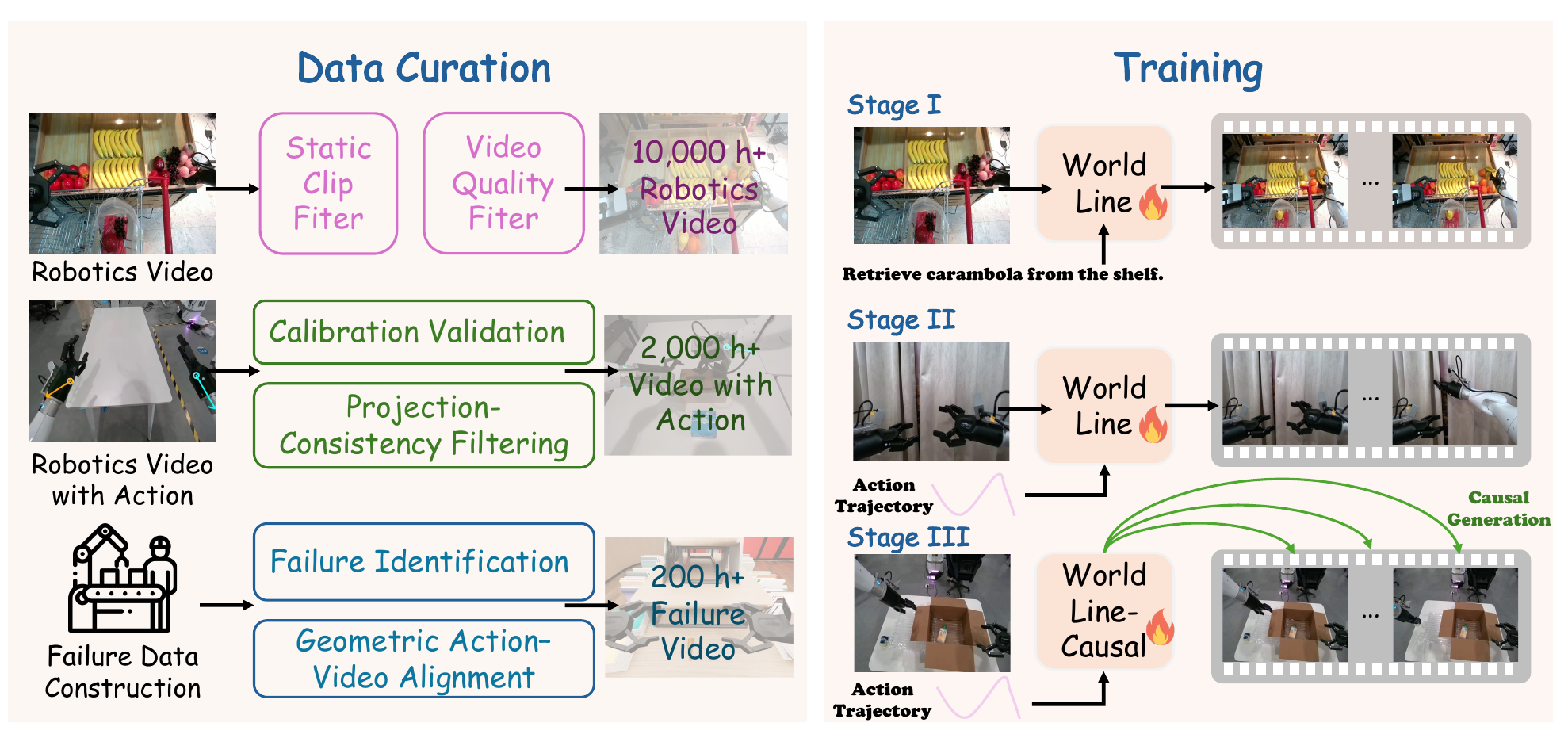}
    \vspace{-0.5cm}
    \caption{\textbf{Data construction and three-stage training of WorldLine.}
Left: action-free videos are filtered for dynamics pretraining; action-labeled
trajectories undergo calibration and projection checks; and failure trajectories
are identified and paired with geometrically aligned controls. Right: Stage I
learns manipulation dynamics, Stage II grounds heterogeneous controls with
multi-view and failure-enriched training, and Stage III distills the model for
efficient causal rollout.}
    \label{fig:method}
\end{figure}

Given initial observations $\mathbf{x}_0^{1:V}$ from $V$ views and a
sequence of robot controls $\mathbf{q}_{1:T}$, WorldLine maps the controls
to camera-aligned action maps
$\mathbf{A}_{1:T}^{1:V}=\mathcal{G}(\mathbf{q}_{1:T})$ and predicts future
observations:
\begin{equation}
    \hat{\mathbf{x}}_{1:T}^{1:V}
    \sim
    p_{\theta}\!\left(
        \cdot
    \mid
    \mathbf{x}_0^{1:V},
    \mathbf{A}_{1:T}^{1:V}
    \right).
\end{equation}
As illustrated in Figure~\ref{fig:method}, WorldLine is trained in three stages.
Stage I learns transferable manipulation dynamics from large-scale action-free
robot videos. Stage II grounds heterogeneous controls in image space while
promoting interaction coherence through multi-view, failure-enriched, and
relational training. Stage III distills the model for efficient causal rollout.

\subsection{Stage I: Robot Dynamics Pretraining}

We initialize WorldLine from a pretrained Cosmos3-Nano video model and adapt it
to the action-free robot video corpus described in
Section~\ref{sec:data_pipeline}. Given future video latents
$\mathbf{z}=\mathbf{z}_{1:T}$, an initial-frame latent $\mathbf{z}_0$, and task
text $y$, we optimize the standard flow-matching objective
\begin{equation}
    \mathcal{L}_{\mathrm{pre}}
    =
    \mathbb{E}_{\mathbf{z},\boldsymbol{\epsilon},\sigma}
    \left[
    \left\|
    \mathbf{v}_{\theta}
    \left(
    \mathbf{z}_{\sigma},
    \sigma
    \mid
    \mathbf{z}_0,
    \mathbf{e}(y)
    \right)
    -(\boldsymbol{\epsilon}-\mathbf{z})
    \right\|_2^2
    \right],
    \qquad
    \mathbf{z}_{\sigma}
    =(1-\sigma)\mathbf{z}+\sigma\boldsymbol{\epsilon},
\end{equation}
where $\boldsymbol{\epsilon}$ is Gaussian noise, $\sigma$ the noise level,
and $\mathbf{e}(y)$ the task-text embedding. Stage I learns a transferable prior
over robot--object dynamics without action supervision to initialize Stage II.
Model initialization and Stage-I training details are provided in
Appendix~\ref{app:stage1_implementation}.

\subsection{Stage II: Action-Grounded Post-Training}

Stage I learns manipulation dynamics but does not link robot controls to future
observations. Stage II converts this prior into an
action-grounded simulator using geometrically verified action--video
trajectories. We remove task-text conditioning and condition generation only on
initial observations and control sequences. Stage-II training
configuration is provided in Appendix~\ref{app:stage2_implementation}.

\subsubsection{Cross-Embodiment Spatial Action Conditioning}

Native robot controls are not directly aligned across embodiments because they
differ in dimensionality, coordinate conventions, and kinematics. For each
control $\mathbf{q}_t$, we use embodiment-specific kinematics to obtain the
commanded end-effector pose and transform it into each camera frame. Let
$\mathbf{u}_{t}^{v}=\pi_v(\mathbf{p}_t)$ denote the projection of its 3D position
$\mathbf{p}_t$ into view $v$. We construct a Gaussian heatmap with fixed
bandwidth $\sigma_g$ centered at $\mathbf{u}_{t}^{v}$:
\begin{equation}
    \mathbf{h}_{t}^{v}(\mathbf{u})
    =
    \exp\!\left(
    -\frac{\|\mathbf{u}-\mathbf{u}_{t}^{v}\|_2^2}{2\sigma_g^2}
    \right).
\end{equation}
Using the same heatmap, we spatially encode the corresponding depth $d_t^v$,
Rot6D orientation $\mathbf{r}_t^v\in\mathbb{R}^{6}$, and gripper state $g_t$.
The resulting nine-channel action map at latent resolution
$H_\ell\times W_\ell$ is
\begin{equation}
    \begin{aligned}
    \mathbf{A}_{t}^{v}
    &=
    \operatorname{Concat}\!\left(
    \mathbf{h}_{t}^{v},
    d_t^v\mathbf{h}_{t}^{v},
    \{r_{t,k}^v\mathbf{h}_{t}^{v}\}_{k=1}^{6},
    g_t\mathbf{h}_{t}^{v}
        \right)
        \in\mathbb{R}^{9\times H_{\ell}\times W_{\ell}},
        \qquad
        \mathbf{A}^{v}
        =\{\mathbf{A}_{t}^{v}\}_{t=1}^{T}.
    \end{aligned}
\end{equation}
The action maps depend only on the control sequence, robot kinematics, and
camera calibration, without using future observations. We encode them with a
lightweight residual branch and add the resulting tokens to the video-latent
embeddings:
\begin{equation}
    \widetilde{\mathbf{A}}^{v}
    =
    \mathbf{A}^{v}
    +
    \operatorname{Conv3D}(\mathbf{A}^{v}),
    \qquad
    \mathbf{E}^{v}
    =
    \mathbf{E}_{\mathrm{video}}^{v}
    +
    \mathbf{W}_{\mathrm{act}}
    \operatorname{Patchify}(\widetilde{\mathbf{A}}^{v}).
\end{equation}
Because the action and video tokens share the same patch layout, their spatial
and temporal coordinates remain aligned. Representing controls through their
camera-space effects provides a shared conditioning interface while preserving
action geometry across embodiments. Implementation details for the action maps
are provided in Appendix~\ref{app:stage2_implementation}.

\subsubsection{Multi-View and Failure-Enriched Training}

Action grounding does not ensure coherent interactions: a single camera
may miss gripper--object contact and object motion, while success-heavy data
underrepresent failure outcomes. For multi-view training, we use trajectories
with synchronized head, left-wrist, and right-wrist views; trajectories missing
a required wrist view are excluded from this objective. The views share temporal
positions but retain view-specific spatial coordinates through RoPE. Each view
receives camera-projected action maps, and the multi-view flow-matching loss is denoted by
$\mathcal{L}_{\mathrm{FM}}^{\mathrm{mv}}$. We apply the same objective to
successful and failed trajectories without outcome labels, exposing the model
to unsuccessful interaction outcomes. Further construction details are provided
in Appendix~\ref{app:data_construction}.

\subsubsection{Relational Dynamics Regularization}

Multi-view and failure-enriched data broaden interaction coverage, but the
flow-matching objective does not explicitly preserve relational dynamics across
time and viewpoints~\cite{PhysisForcing}. We therefore regularize intermediate
DiT features using a frozen V-JEPA2~\cite{vjepa} applied to the target
videos. Let $\mathbf{S}_{t}^{v}$ and $\mathbf{T}_{t}^{v}$ denote the student and
teacher feature tokens at time $t$ and view $v$. For row-normalized token features, their pairwise cosine-similarity matrix is
$\mathcal{R}(\mathbf{F},\mathbf{G})
=\overline{\mathbf{F}}\,\overline{\mathbf{G}}^{\top}$.

We match student and teacher relations between consecutive frames within each
view and between synchronized frames across views:
\begin{equation}
\begin{aligned}
    \mathcal{L}_{\mathrm{intra}}
    &=
    \mathbb{E}_{v,t}
    \left[
    \left\|
    \mathcal{R}(\mathbf{S}_{t}^{v},\mathbf{S}_{t+1}^{v})
    -
    \mathcal{R}(\mathbf{T}_{t}^{v},\mathbf{T}_{t+1}^{v})
    \right\|_1
    \right],\\
    \mathcal{L}_{\mathrm{cross}}
    &=
    \mathbb{E}_{(v,v'),t}
    \left[
    \left\|
    \mathcal{R}(\mathbf{S}_{t}^{v},\mathbf{S}_{t}^{v'})
    -
    \mathcal{R}(\mathbf{T}_{t}^{v},\mathbf{T}_{t}^{v'})
    \right\|_1
    \right].
\end{aligned}
\end{equation}
The complete action-grounding objective is
\begin{equation}
    \mathcal{L}_{\mathrm{post}}
    =
    \mathcal{L}_{\mathrm{FM}}^{\mathrm{mv}}
    +
    \lambda_{\mathrm{intra}}^{(k)}\mathcal{L}_{\mathrm{intra}}
    +
    \lambda_{\mathrm{cross}}^{(k)}\mathcal{L}_{\mathrm{cross}}.
\end{equation}
At optimization step $k$, the weights are dynamically scaled relative to the flow-matching loss; the
EMA-based schedule and target ratios are provided in
Appendix~\ref{app:relational_regularization}. This transfers the teacher's
temporal and cross-view relational structure to WorldLine.

\subsection{Stage III: Robot-Focused Causal Distillation}

Stage II generates full trajectories from a fixed action sequence through
iterative ODE sampling, preventing online control updates and making repeated
rollout costly. We distill it into a block-autoregressive student that accepts
new controls at each block and generates the future in four denoising steps.
For block $b$, $\mathbf{h}_b$ denotes the available causal history and
$\mathbf{A}_b$ the action maps for the next block, which spans four latent
frames (16 RGB frames). Training proceeds through causal adaptation, trajectory
regression, and score-based distribution matching~\cite{dmd2}.

Few-step distillation can overemphasize static backgrounds~\cite{zhu2026causal}. Let
$\overline{\mathbf{M}}$ denote the latent-resolution robot mask and
$\widehat{\mathbf{z}}$ and $\mathbf{z}_{T}$ the student and teacher latents. We
define
\begin{equation}
\mathcal{L}_{\mathrm{RF}}
=
    \lambda_{\mathrm{robot}}
    \left\|\overline{\mathbf{M}}\odot(\widehat{\mathbf{z}}-\mathbf{z}_{T})\right\|_2^2
    +
    \lambda_{\Delta\mathrm{robot}}
\left\|\overline{\mathbf{M}}\odot
(\Delta_t\widehat{\mathbf{z}}-\Delta_t\mathbf{z}_{T})\right\|_1.
\end{equation}
These are the weighted robot-reconstruction and temporal-motion auxiliaries in
$\mathcal{L}_{\mathrm{SF\text{-}DMD}}$, denoted by
$\mathcal{L}_{\mathrm{robot}}$ and $\mathcal{L}_{\Delta\mathrm{robot}}$ in the
Appendix; trajectory regression is a separate objective. The model enables
low-latency causal rollout, online updates, and
mask-free inference. Distillation stages and inference settings are detailed in
Appendix~\ref{app:causal_distillation} and Appendix~\ref{app:inference_configuration}.

\section{Experiments}

We evaluate WorldLine's capabilities through three sets of experiments:
action-conditioned prediction across in- and out-of-domain settings, policy
evaluation, and embodied planning. We additionally assess causal rollout
efficiency and ablate the key data, representation, and training components.

\subsection{Action-Conditioned Video Generation}

\noindent\textbf{Evaluation protocol.}
Given an initial observation and recorded action sequence, each model predicts
the future video. We evaluate on scene-disjoint
AgiBotWorld~\cite{Agibotworld} trajectories with successful and failed
executions, and on DROID~\cite{droid}, which is excluded from WorldLine
training, adaptation, and checkpoint selection. The settings test
generalization to unseen in-domain scenes and fully out-of-domain environments
and robot embodiments, respectively.

\noindent\textbf{Metrics.}
We report PSNR and SSIM for pixel fidelity and LPIPS for perceptual
similarity. To address static backgrounds, we report head-view robot-mask IoU
for robot motion. SAM~3~\cite{sam3} independently segments both robot masks;
projected end-effector location only initializes segmentation. Thus, IoU
measures image-space motion agreement. We evaluate on held-out AgiBotWorld and
out-of-domain DROID; details are in
Appendix~\ref{app:action_generation_evaluation}.

\noindent\textbf{Quantitative results.}
Table~\ref{tab:action_conditioned_generation} compares WorldLine with video
generation and action-conditioned simulation baselines. On successful
AgiBotWorld trajectories, WorldLine ranks first in SSIM, LPIPS, and robot IoU
and second in PSNR. On failures, it leads all four metrics and improves robot
IoU by 0.1626 over the strongest baseline. Although DROID is out-of-domain for
WorldLine and in-domain for Ctrl-World and Masked Visual Actions, WorldLine
achieves the highest robot IoU (0.2739), exceeding the best prior result by
0.0690, with near-best visual quality. Its causal variant ranks second on
failure metrics and achieves the best DROID LPIPS.

\begin{figure*}[t]
    \centering
    \vspace{-0.3cm}
    \includegraphics[width=\textwidth]{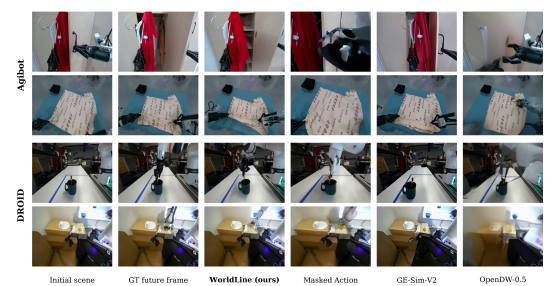}
    \vspace{-0.6cm}
    \caption{\textbf{Qualitative action-conditioned video generation.} Given
initial frame and actions, each model predicts a frame on held-out AgiBotWorld
(top) and out-of-domain DROID (bottom). Causal WorldLine better matches
ground-truth robot motion and gripper--object configuration.}
    \label{fig:qualitative_action_generation}
\end{figure*}

\noindent\textbf{Qualitative results.}
Figure~\ref{fig:qualitative_action_generation} shows how the quantitative gains
translate into more accurate robot motion and resulting object configurations.
Across AgiBotWorld and DROID, causal WorldLine aligns robot motion with the input controls while
preserving the resulting gripper--object configuration. Competing methods
instead exhibit robot drift, mismatched gripper--object configurations, or
insufficient motion,
particularly under DROID's unseen environments and embodiments.

\begin{table*}[t]
    \centering
    \caption{Action-conditioned video prediction on held-out AgiBotWorld
    success/failure trajectories and DROID, averaged over five runs.
    $\dagger$ denotes DROID-trained methods, making DROID in-domain.
    Arrows indicate metric direction; green and blue mark the best and second-best results.}
    \label{tab:action_conditioned_generation}
    \vspace{0.2cm}
    \scriptsize
    \setlength{\tabcolsep}{2.8pt}
    \resizebox{\textwidth}{!}{%
    \begin{tabular}{@{}lcccccccccccc@{}}
        \toprule
        & \multicolumn{8}{c}{\textbf{In-Domain}}
        & \multicolumn{4}{c}{\textbf{DROID Evaluation}} \\
        \cmidrule(lr){2-9}\cmidrule(lr){10-13}
        \textbf{Model}
        & \multicolumn{4}{c}{\textbf{AgiBotWorld (Successful)}}
        & \multicolumn{4}{c}{\textbf{AgiBotWorld (Failure)}}
        & \multicolumn{4}{c}{\textbf{DROID}} \\
        \cmidrule(lr){2-5}\cmidrule(lr){6-9}\cmidrule(lr){10-13}
        & PSNR$\uparrow$ & SSIM$\uparrow$ & LPIPS$\downarrow$ & Robot IoU$\uparrow$
          & PSNR$\uparrow$ & SSIM$\uparrow$ & LPIPS$\downarrow$ & Robot IoU$\uparrow$
          & PSNR$\uparrow$ & SSIM$\uparrow$ & LPIPS$\downarrow$ & Robot IoU$\uparrow$ \\
        \midrule
        \rowcolor{LightOrange}
        \multicolumn{13}{c}{\textbf{Video Generation Models}} \\
        Cosmos3-Nano~\cite{cosmos3} 
          & 12.54 & 0.4752 & 0.5654 & 0.2804 & 10.35 & 0.4723 & 0.6000 & 0.1444 & 16.97 & 0.6516 & 0.3492 & \cellcolor{BestInModule}\underline{0.2049} \\ 
        Cosmos3-Super~\cite{cosmos3}   
          & 12.57 & 0.4685 & 0.5674 & 0.2975 & 10.73 & 0.4786 & 0.6216 & 0.1755 & 17.07 & 0.6524 & 0.3457 & 0.1978 \\
        \rowcolor{LightOrange}
        \multicolumn{13}{c}{\textbf{Action-Conditioned Models}} \\
        Ctrl-World$^{\dagger}$ ~\cite{ctrl-world} 		
          & 15.68 & 0.6457 & 0.4690 & 0.2863 & 14.73 & 0.6881 & 0.4140 & 0.1917 & \cellcolor{BestOverall}\textbf{19.11} & \cellcolor{BestOverall}\textbf{0.7502} & 0.3010 & 0.1686 \\
        GE-Sim-V2 ~\cite{gesim} 
          & \cellcolor{BestOverall}\textbf{18.90} & \cellcolor{BestInModule}\underline{0.7334} & \cellcolor{BestInModule}\underline{0.2803} & \cellcolor{BestInModule}\underline{0.6300}
          & 15.67 & 0.7160 & 0.3285 & 0.3079
          & 16.97 & 0.6771 & 0.3766 & 0.1959 \\
        OpenDW-0.5~\cite{opendw}
          & 16.67 & 0.6681 & 0.4639 & 0.4821 & 13.10 & 0.6364 & 0.5204 & 0.2187 & 15.53 & 0.6247 & 0.5012 & 0.0827 \\
        Masked Visual Actions$^{\dagger}$ ~\cite{maskedvisual} 
          & 14.53 & 0.6378 & 0.4208 & 0.3649 & 13.48 & 0.7143 & 0.3414 & 0.2083 & 17.05 & 0.7104 & 0.3635 & 0.1965 \\
        \rowcolor{LightOrange}
        \multicolumn{13}{c}{\textbf{Ours}} \\
        \textbf{WorldLine}
          & \cellcolor{BestInModule}\underline{18.67} & \cellcolor{BestOverall}\textbf{0.7931} & \cellcolor{BestOverall}\textbf{0.1695} & \cellcolor{BestOverall}\textbf{0.6539}
          & \cellcolor{BestOverall}\textbf{19.19} & \cellcolor{BestOverall}\textbf{0.8135} & \cellcolor{BestOverall}\textbf{0.1913} & \cellcolor{BestOverall}\textbf{0.4705}
          & \cellcolor{BestInModule}\underline{19.04} & \cellcolor{BestInModule}\underline{0.7480} & \cellcolor{BestInModule}\underline{0.2968} & \cellcolor{BestOverall}\textbf{0.2739} \\
        \textbf{WorldLine (Causal 4-Step)}   
          & 18.31 & 0.7206 & 0.2876 & 0.5923 & \cellcolor{BestInModule}\underline{18.32} & \cellcolor{BestInModule}\underline{0.7910} & \cellcolor{BestInModule}\underline{0.2241} & \cellcolor{BestInModule}\underline{0.4265} & 18.90 & 0.7232 & \cellcolor{BestOverall}\textbf{0.2705} & 0.1986 \\
        \bottomrule
    \end{tabular}%
    }
\end{table*}

\begin{figure*}[t]
    \centering
    \includegraphics[width=\textwidth]{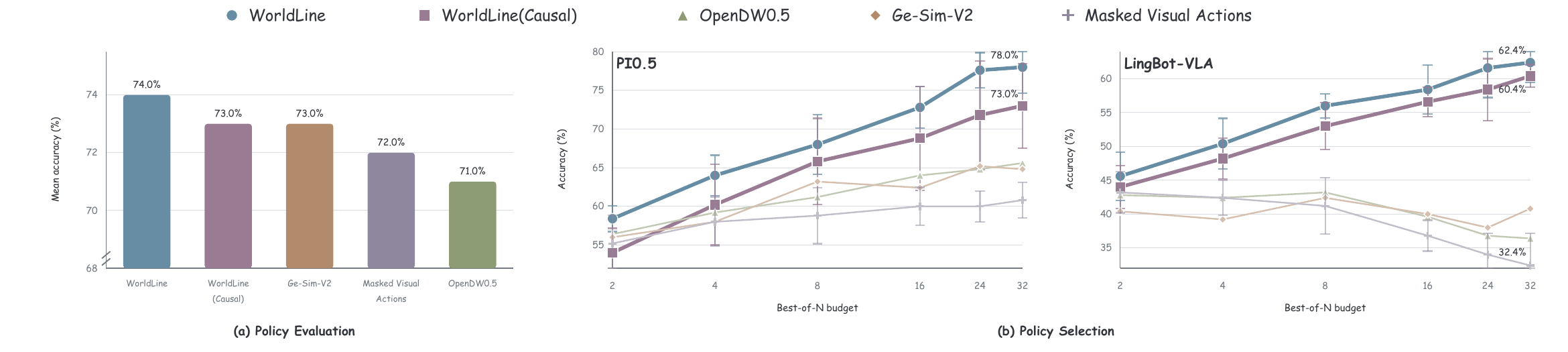}
    \vspace{-0.5cm}
    \caption{\textbf{WorldLine for policy evaluation and selection.}
    (a) Mean policy-evaluation accuracy over five generation seeds on RoboTwin
    and AgiBot. (b) RoboTwin task success rate under different rollout budgets
    for $\pi_{0.5}$ and LingBot-VLA.}
    \label{fig:downstream_evaluation}
\end{figure*}
\begin{figure*}[t]
    \centering
    \includegraphics[height=0.135\textheight]{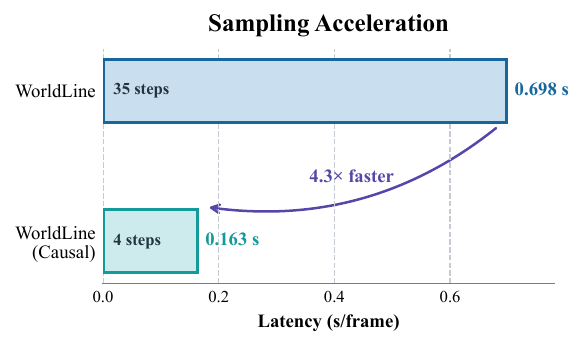}
    \hfill
    \includegraphics[height=0.135\textheight]{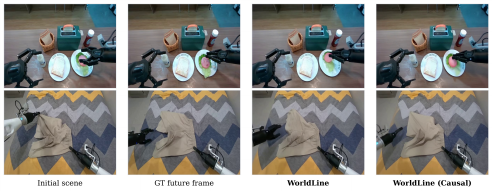}
    \vspace{-0.3cm}
    \caption{\textbf{Causal distillation efficiency and quality.}
Left: sampling steps and per-frame latency under identical settings. Right:
Stage-II and causal predictions on AgiBot. The causal model accelerates rollout
while preserving appearance, robot geometry, and object evolution.}
    \label{fig:causal_efficiency}
\end{figure*}

\subsection{Downstream Robotic Applications}

Beyond video generation, a visual simulator must capture task-relevant action
consequences to support decisions; we therefore evaluate
WorldLine on policy evaluation and embodied planning.

\subsubsection{Policy Evaluation}

\noindent\textbf{Evaluation protocol.}
We evaluate task-success prediction from rollouts on RoboTwin and AgiBot. From
an initial observation and action sequence, each simulator generates a video
that Qwen3VL-8B classifies as successful or unsuccessful. Accuracy is averaged
over five generation seeds.

\noindent\textbf{Results.}
Figure~\ref{fig:downstream_evaluation}(a) shows that WorldLine achieves the
highest mean accuracy of 74\%, followed by causal WorldLine and GE-Sim-V2 at
73\%, Masked Visual Actions at 72\%, and OpenDW-0.5 at 71\%. Causal WorldLine
remains within one point of Stage II at lower latency. Evaluation sets,
evidence-extraction prompts, and accuracy computation are described in
Appendix~\ref{app:policy_evaluation}.

\subsubsection{Embodied Planning}

\noindent\textbf{Evaluation protocol.}
We evaluate out-of-domain rollout-based planning on RoboTwin~\cite{robotwin},
which is excluded from WorldLine training, adaptation, and checkpoint
selection. Across 10 tasks with five scenes each, $\pi_{0.5}$~\cite{pi0.5} and
LingBot-VLA~\cite{lingbot-vla} each sample
$N\in\{2,4,8,16,24,32\}$ candidate action trajectories. Each simulator predicts
their outcomes, Qwen3VL-8B scores task completion, and the highest-scoring
trajectory is executed.

\noindent\textbf{Baselines and metrics.}
We compare Stage-II and causal WorldLine with GE-Sim-V2, OpenDW-0.5, and
Masked Visual Actions using shared candidates and Qwen3VL-8B~\cite{qwen3vl}
selector. We report per-budget success and direct-execution gains over five
independently sampled banks. Candidate generation, the selection
procedure, and configuration are detailed in
Appendix~\ref{app:embodied_planning}.

\noindent\textbf{Results.}
WorldLine benefits as rollout budgets grow, while prior simulators
scale weakly; because Qwen3VL-8B is imperfect, candidates do not guarantee
monotonic gains. At $N=32$, selection benefits both policies and narrows their
direct-execution success gap from 17.9 to 15.6 percentage points. This suggests
the simulator can improve action selection across policy proposals, not just one
policy's distribution. Four-step causal rollouts retain most gains at lower
cost, preserving outcome distinctions for ranking. They persist without
WorldLine training or adaptation on RoboTwin.

\subsection{Causal Few-Step Acceleration}

\noindent\textbf{Evaluation protocol.}
We compare Stage-II and four-step causal WorldLine at batch size one under
identical 129-frame settings to isolate speedup, reporting sampling steps and
per-frame latency.

\noindent\textbf{Results.}
Causal distillation reduces sampling from 35 to four steps, cutting 129-frame
generation from 90 to 21 seconds and per-frame latency from 0.698 to 0.163
seconds, a $4.3\times$ speedup. Figure~\ref{fig:causal_efficiency} shows that
this acceleration preserves the main robot--object evolution in two AgiBot
examples. Hardware, timing, and rollout settings are provided in
Appendix~\ref{app:causal_distillation} and
Appendix~\ref{app:inference_configuration}.

\noindent\textbf{Analysis.}
Policy evaluation asks whether a rollout conveys task completion, whereas
planning must preserve relative distinctions among competing futures. A
prediction can therefore retain enough evidence for binary success assessment
yet still reverse the ranking of similar candidates. The appendix results
suggest this sensitivity depends on the task: few-step rollouts remain useful
when outcomes are visually distinguishable, but are less reliable when
selection depends on finer differences. Thus, acceleration does not impose a
uniform quality loss; it changes which task-relevant cues survive in the
rollout. Since the candidate budget is fixed, whether using the saved runtime
to evaluate more candidates can offset ranking errors remains an open question.

\begin{figure*}[t]
    \centering
    \includegraphics[width=\textwidth]{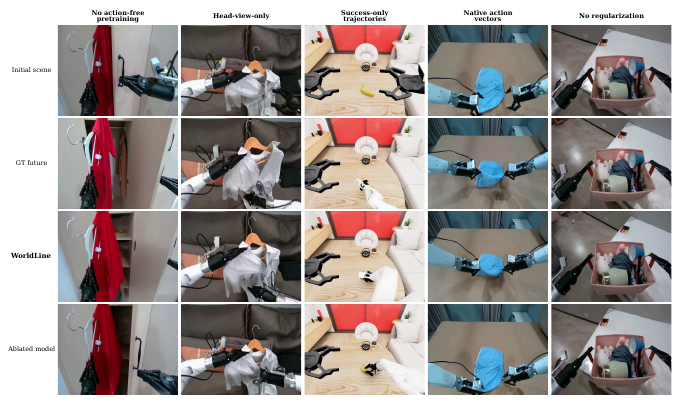}
    \vspace{-0.7cm}
    \caption{\textbf{Qualitative comparison of WorldLine ablations.}
    We compare the full model with variants that remove key data,
    representation, and training components, highlighting their effects on
    robot motion and robot--object interaction outcomes.}
    \label{fig:qualitative_ablation}
\end{figure*}

\begin{table*}[t]
    \centering
    \caption{\textbf{Ablation of WorldLine components.} LPIPS and robot-mask IoU are
    evaluated on held-out AgiBotWorld and out-of-domain DROID. RoboTwin planning
    uses 32 rollouts; gains are over direct execution. LPIPS is omitted for
    head-view-only. The no-action-free variant matches the full model's
    training-compute budget. Green and blue mark best and second-best results.}
    \vspace{0.2cm}
    \label{tab:ablation_studies}
    \scriptsize
    \setlength{\tabcolsep}{3.0pt}
    \resizebox{\textwidth}{!}{%
    \begin{tabular}{@{}lcccccccc@{}}
        \toprule
        & \multicolumn{2}{c}{\textbf{AgiBotWorld}}
        & \multicolumn{2}{c}{\textbf{DROID}}
        & \multicolumn{2}{c}{\textbf{RoboTwin: $\pi_{0.5}$}}
        & \multicolumn{2}{c}{\textbf{RoboTwin: LingBot-VLA}} \\
        \cmidrule(lr){2-3}\cmidrule(lr){4-5}
        \cmidrule(lr){6-7}\cmidrule(lr){8-9}
        \textbf{Variant}
        & LPIPS$\downarrow$ & Robot IoU$\uparrow$
        & LPIPS$\downarrow$ & Robot IoU$\uparrow$
        & Success$\uparrow$ & Gain$\uparrow$
        & Success$\uparrow$ & Gain$\uparrow$ \\
        \midrule
        \textbf{Full WorldLine}
        & \cellcolor{BestOverall}\textbf{0.1804}
        & \cellcolor{BestOverall}\textbf{0.5622}
        & \cellcolor{BestOverall}\textbf{0.2968}
        & \cellcolor{BestOverall}\textbf{0.2739}
        & \cellcolor{BestOverall}\textbf{78.0±1.4\%}
        & \cellcolor{BestOverall}\textbf{+19.1\%}
        & \cellcolor{BestOverall}\textbf{62.4±3.6\%}
        & \cellcolor{BestOverall}\textbf{+21.4\%} \\
        \rowcolor{LightOrange}
        \multicolumn{9}{c}{\textbf{Data}} \\
        \quad No action-free data (compute-matched)
        & 0.2924
        & \cellcolor{BestInModule}\underline{0.5277}
        & \cellcolor{BestInModule}\underline{0.3028}
        & 0.2614 & 73.2±3.3\% & +14.3\% & 60.4±3.0\% & +19.4\% \\
        \quad Head-view-only data
        & -- & 0.4348 & --
        & \cellcolor{BestInModule}\underline{0.2665}
        & 72.8±1.10 \% & +13.9\%
        & \cellcolor{BestInModule}\underline{62.0 ±3.2\%}
        & \cellcolor{BestInModule}\underline{+21.0\%} \\ 
        \quad Success-only trajectories
        & \cellcolor{BestInModule}\underline{0.2401}
        & 0.4721
        & \cellcolor{BestInModule}\underline{0.3028}
        & 0.2566
        & \cellcolor{BestInModule}\underline{77.2±2.3\%}
        & \cellcolor{BestInModule}\underline{+18.3\%}
        & 51.6±3.0\% & +10.6\% \\
        \rowcolor{LightOrange}
        \multicolumn{9}{c}{\textbf{Action Control}} \\ 
        \quad Native action vectors
        & 0.3118 & 0.3639 & 0.4537 & 0.1171 & 67.2±2.3\% & +8.3\% & 53.2±3.3\% & +12.2\% \\
        \rowcolor{LightOrange}
        \multicolumn{9}{c}{\textbf{Training Strategy}} \\
        \quad No regularization
        & 0.2477 & 0.4593 & 0.3386 & 0.2303
        & \cellcolor{BestInModule}\underline{77.2±3.0\%}
        & \cellcolor{BestInModule}\underline{+18.3\%}
        & \cellcolor{BestInModule}\underline{62.0±2.4\%}
        & \cellcolor{BestInModule}\underline{+21.0\%} \\
        \bottomrule
    \end{tabular}%
    }
\end{table*}

\subsection{Ablation Studies}

\noindent\textbf{Evaluation protocol.}
We evaluate each variant using LPIPS and robot-mask IoU on held-out AgiBotWorld
and DROID, plus RoboTwin planning success. For planning, Qwen3VL-8B selects
from the same 32 trajectories sampled by $\pi_{0.5}$ and LingBot-VLA for every
variant, matching the $N=32$ setting in the main planning experiment.
The variant without action-free data excludes the Stage-I corpus but uses the
same total training-compute budget as the full model. The complete ablation
protocol is provided in Appendix~\ref{app:ablation_protocol}.

\noindent\textbf{Data.}
Under matched total training compute, Table~\ref{tab:ablation_studies} shows
that removing action-free data degrades visual quality and planning. Because
the compute budget is matched, this points to the corpus's complementary
interaction coverage rather than additional optimization. The visual degradation
is more pronounced on held-out AgiBotWorld than on DROID, suggesting that the
action-free corpus contributes most directly to visual fidelity in the familiar
domain while still supporting robot-motion prediction under domain shift.
Head-view-only training further reduces robot-mask overlap, particularly on
AgiBotWorld, consistent with wrist views adding close-range contact and
object-motion cues. Failure trajectories broaden the observed action-conditioned
outcomes beyond successful task completion. Their benefit is policy-dependent:
success-only data barely changes $\pi_{0.5}$ performance but lowers LingBot-VLA
success by 10.8 points. Since candidate sets are fixed across variants, this
suggests that failure coverage changes how well predicted outcomes rank policy
proposals, rather than uniformly increasing success across policies.

\begin{wrapfigure}{r}{0.42\columnwidth}
    \centering
    \IfFileExists{Figures/action_encoding_scaling.pdf}{%
        \includegraphics[width=\linewidth]{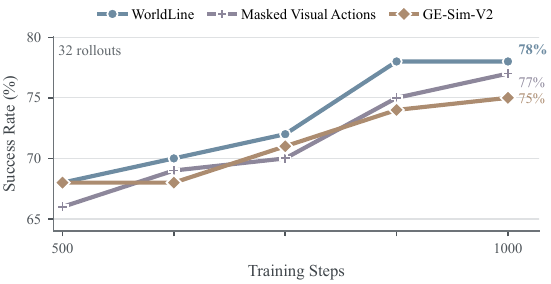}%
    }{%
        \fbox{\parbox[c][0.065\textheight][c]{0.90\linewidth}{
            \centering
            \textbf{Action-Encoding Scaling}\\[0.3em]
            Planning success vs. budget
        }}%
    }
    \vspace{-0.3cm}
    \caption{Matched action-encoding comparison on RoboTwin.}
    \label{fig:action_encoding_scaling}
    \vspace{-1.0\baselineskip}
\end{wrapfigure}

\noindent\textbf{Action encoding and grounding.}
Replacing image-space action maps with FiLM-injected native action vectors
reduces robot IoU, worsens LPIPS, and lowers planning success across both
datasets and policies (Table~\ref{tab:ablation_studies}). Unlike camera-space
maps, native vectors do not explicitly associate each control with its projected
image location and its resulting visual effect. The consistent degradation therefore suggests that spatial
grounding helps the model predict not only plausible motion but also the
consequences used to rank candidate actions. Among projection-based encodings, WorldLine
outperforms GE-Sim-V2 and Masked Visual Actions under matched RoboTwin training
and a 32-rollout budget as shown in Figure~\ref{fig:action_encoding_scaling}.
This advantage across both policies indicates that the benefit is not specific
to one policy's action prior.

\noindent\textbf{Training strategy.}
Removing relational regularization worsens visual quality and robot-mask IoU,
yet has little effect on planning success (Table~\ref{tab:ablation_studies}).
This contrast suggests that the regularizer primarily improves temporal and
cross-view geometric consistency across the full generated sequence. Under the fixed-candidate evaluation, such
improvements need not change which action proposal is ranked highest, even when
they make the predicted robot motion more faithful. Figure~\ref{fig:qualitative_ablation}
illustrates these differences in robot motion and interaction outcomes.

\section{Future Work}

Future work will extend WorldLine to rare manipulation patterns, deformable
objects, diverse cameras, and longer action-conditioned horizons. These settings
can test whether image-space action grounding and cross-embodiment dynamics
remain reliable under greater variation. We will study WorldLine as an RL
environment for training robot policies and as a test-time scaling tool,
allocating more candidate rollouts to difficult tasks. Better uncertainty
estimates and closed-loop replanning may help planners identify unreliable
predictions, allocate computation efficiently, and test whether improved
simulated outcomes translate into more reliable real-robot behavior.

\section{Conclusion}

We presented WorldLine, an action-driven visual simulator that scales dynamics
learning beyond scarce action-labeled data by separating action-free pretraining
from image-space action grounding. This design learns transferable
robot--object dynamics while grounding heterogeneous controls across
embodiments and improving action-conditioned robot-motion and outcome prediction.
Across held-out and fully out-of-domain settings, WorldLine follows commanded actions, predicts
trajectory success, and improves embodied planning through rollout selection.
Four-step causal distillation retains these capabilities at substantially lower
latency. Thus, action-driven visual simulation can
scale across embodiments while remaining useful for downstream robotic
decisions.

\bibliography{iclr2027_conference}
\bibliographystyle{iclr2027_conference}

\clearpage

\appendix

\section{Data Construction Details}
\label{app:data_construction}

\subsection{Stage-Wise Data Composition}
\label{app:stage_wise_data}

WorldLine uses complementary data in its first two training stages. Stage I
learns general robot motion and object-interaction dynamics from a diverse
collection of robot videos. For each trajectory, we select a primary
task-facing view, typically a head, top, or front camera that clearly observes
the robot workspace. Although some source datasets contain robot states or
actions, Stage I does not provide these signals to the model and treats all
samples as action-free videos.

Stage II uses robot trajectories with sufficiently reliable state or action
sequences and camera geometry. Unlike Stage I, it focuses on grounding robot
motion through spatial action conditions and jointly processes synchronized
head, left-wrist, and right-wrist observations when all required views are
available.

\subsection{Video Sampling and Motion Filtering}
\label{app:video_sampling}

Training clips are sampled dynamically from valid trajectory segments rather
than being pre-segmented into fixed, non-overlapping clips. The sampled clips
contain between 33 and 129 frames, allowing the model to observe interactions
at different temporal scales. Stage I uses a larger temporal stride to cover
longer motion, whereas Stage II samples more densely to preserve fine-grained
action effects.

Following Figure~\ref{fig:method}, we apply source-specific motion and
video-quality criteria to remove static, corrupted, and
chassis-motion-dominated clips. Videos are resized to source-specific training
resolutions, with camera intrinsics adjusted accordingly.

\subsection{Action Geometry and Multi-View Construction}
\label{app:action_geometry}

For Stage II, we retain trajectories that pass calibration validation and
projection-consistency filtering, as summarized in Figure~\ref{fig:method}.
Visual observations and controls are aligned by frame index, and forward
kinematics transforms each end-effector pose into the corresponding camera
frame. The projected position defines a Gaussian spatial condition that also
encodes depth, Rot6D orientation, and gripper state, producing the nine-channel
action representation described in the main paper. Invalid or out-of-view
projections are masked, and wrist-camera poses are updated from robot
kinematics when dynamic extrinsics are available.

Stage II uses three semantic camera slots corresponding to the head or top
camera, left wrist, and right wrist. All views within a trajectory share the
same sampled frame indices and clip length. Trajectories missing a required
wrist view are excluded from multi-view training rather than automatically
converted into single-view samples.

\subsection{Failure Data and Evaluation Separation}
\label{app:failure_data_evaluation}

\noindent\textbf{Failure trajectory construction.}
In addition to successful demonstrations, we construct a dedicated failure
dataset by collecting controlled robot executions that do not successfully
complete the intended manipulation task. These trajectories capture diverse
unsuccessful interactions, such as missed grasps, object slippage, unintended
collisions, and incomplete manipulation. We retain the corresponding multi-view
observations, robot states or actions, and camera geometry, and process them
using the same temporal sampling and action-grounding pipeline as the remaining
Stage-II data. During training, failure trajectories are treated as ordinary
action--video pairs without providing explicit outcome labels. The simulator
must therefore learn the visual consequences of unsuccessful actions directly
from the observed trajectories rather than from a predefined failure category.

\noindent\textbf{Evaluation separation.}
For evaluation, we construct a scene-disjoint AgiBotWorld split such that the
evaluation scenes and their associated trajectories are excluded from
training. DROID is not used in any WorldLine training stage, model adaptation,
or checkpoint selection. We directly evaluate the standard WorldLine
checkpoint on DROID. Compared with the predominantly bimanual training data,
DROID contains substantially different camera viewpoints and single-arm
manipulation settings, providing an out-of-distribution evaluation of viewpoint
and embodiment-configuration generalization.

\section{Implementation Details}

\subsection{Model Initialization and Stage-I Training}
\label{app:stage1_implementation}

WorldLine is built upon the pretrained Cosmos3-Nano video generation model.
All existing model components are initialized from pretrained weights. Stage I
performs full-parameter training on the action-free robot video corpus. The
model is conditioned on the initial observation and task description and is
optimized using the flow-matching objective described in the main paper. This
stage adapts the pretrained video prior to robot motion and manipulation
dynamics without consuming robot actions.

\subsection{Stage-II Action-Grounded Post-Training}
\label{app:stage2_implementation}

Stage II is initialized from the Stage-I model and removes task-text
conditioning. All existing modules inherit their weights from the preceding
checkpoint, while the newly introduced action-conditioning branch is
zero-initialized. The zero initialization preserves the behavior of the
action-free model at the beginning of post-training and allows action control
to be introduced progressively.

For each view, the nine-channel action maps are rendered directly at the video
latent resolution. For a $480\times640$ input, the corresponding action tensor
has shape $9\times T_{\ell}\times30\times40$. We use a fixed Gaussian bandwidth
of $\sigma_g=10$ pixels on the $384\times512$ reference canvas. The action
encoder applies a single $\operatorname{Conv3D}(9,9)$ layer with a
$1\times3\times3$ kernel and adds its output to the input action maps. The
resulting representation is spatially patchified using $2\times2$ patches and
projected from 36 to 4096 dimensions. The action tokens are then added to the
spatially and temporally aligned video tokens before the first DiT block. No
numerical-action FiLM modulation is used.

The three camera views are ordered as head, left wrist, and right wrist. They
share the same temporal coordinates, while their RoPE height coordinates begin
at fixed semantic slots of 0, 480, and 992, respectively. During Stage II, we
optimize the generation-specific MoE modules, time embeddings, video-latent
input and output projections, and the action-conditioning branch, while keeping
the remaining pretrained components fixed. We use AdamW with betas
$(0.9,0.95)$, weight decay $10^{-6}$, and a base learning rate of
$1.5\times10^{-5}$. The newly initialized action branch uses a 66-times
learning-rate multiplier. Training uses a per-device batch size of one,
gradient accumulation of two, and gradient clipping at 0.2.

\subsection{Relational Regularization}
\label{app:relational_regularization}

We use a frozen V-JEPA2 ViT-g/384 encoder as the relational teacher. We choose
V-JEPA2 because its self-supervised representations capture semantic motion and
interaction structure beyond low-level pixel similarity, providing stable
targets for modeling robot--object dynamics without requiring additional
annotations.

Teacher features are obtained by concatenating the normalized outputs of its
last four Transformer blocks. Student features are extracted from the output
of the 18th DiT block. Because the teacher and student have different feature
dimensions, we do not directly regress their feature values. Instead, both
representations are temporally aggregated into eight bins and spatially pooled
to an $8\times8$ grid. Relational supervision is then applied to their
cosine-similarity matrices.

The intra-view objective aligns token relations between adjacent temporal bins
within the same camera view. At each optimization step, one of the head,
left-wrist, and right-wrist views is selected in a round-robin manner. The
cross-view objective aligns the head view with one wrist view and alternates
between the left and right wrists.

Rather than using fixed loss coefficients, we dynamically scale the relational
objectives according to exponential moving averages of their ratios to the
flow-matching loss. The target ratios are 0.03 for intra-view regularization and
0.005 for cross-view regularization, with maximum scales of 0.20 and 0.10,
respectively. The scales are gradually introduced during the first 250
optimization steps. The V-JEPA2 teacher remains frozen throughout training.

\subsection{Causal Distillation and Rollout}
\label{app:causal_distillation}

We convert the non-causal Stage-II model into an efficient block-autoregressive
simulator through three successive stages: causal teacher-forcing adaptation,
four-step trajectory regression, and score-based distribution matching. Each
stage is initialized from the checkpoint produced by the preceding stage.

\subsubsection{Causal Teacher-Forcing Adaptation}

We first introduce causal attention and block-wise generation while retaining
the Stage-II flow-matching objective. The model receives one initial latent
frame as the observation condition and generates four latent frames per block.
Given the temporal compression ratio of the video tokenizer, each generated
block corresponds to 16 RGB frames. During training, each block is conditioned
on ground-truth preceding blocks and the action maps covering the same temporal
interval. This stage teaches the model to generate future blocks under a causal
attention pattern while preserving the action-grounded dynamics learned by the
Stage-II model.

\subsubsection{Four-Step Trajectory Regression}

We next use the frozen causal teacher to construct offline four-step denoising
trajectories. Each block is generated using four explicit Euler updates with
the noise schedule
\begin{equation}
    1.0 \rightarrow \frac{15}{16} \rightarrow \frac{5}{6}
    \rightarrow \frac{5}{8} \rightarrow 0.
\end{equation}
At each noisy node, the student regresses the next node on the teacher
trajectory. An additional terminal objective preserves the final clean
prediction:
\begin{equation}
    \mathcal{L}_{\mathrm{traj}}
    =
    \lambda_{\mathrm{next}}
    \sum_{k=0}^{K-1}
    \left\|
    \widehat{\mathbf{z}}^{(k+1)}-\mathbf{z}_{T}^{(k+1)}
    \right\|_2^2
    +
    \lambda_{\mathrm{terminal}}
    \left\|
    \widehat{\mathbf{z}}^{\mathrm{clean}}
    -\mathbf{z}_{T}^{\mathrm{clean}}
    \right\|_2^2
    +\mathcal{L}_{\mathrm{aux}},
\end{equation}
where $\mathbf{z}_{T}$ denotes the trajectory generated by the frozen teacher,
$\lambda_{\mathrm{next}}=1.0$, and
$\lambda_{\mathrm{terminal}}=0.5$. This stage compresses the teacher's
iterative denoising trajectory into four student updates.

\subsubsection{Score-Based Distribution Matching}

Trajectory regression encourages the student to follow individual teacher
trajectories but does not directly align the overall generated distribution~\cite{selfforcing,zhu2026causal}.
We therefore further optimize the student using score-based distribution
matching~\cite{dmd,dmd2}. Given a student-generated clean latent $\mathbf{G}_{\theta}$, we add
noise to obtain $\mathbf{x}_t$. A frozen teacher estimates the flow field of the
real data distribution, while a trainable fake-score model estimates the flow
field of the generated distribution:
\begin{equation}
    \mathbf{g}_{\mathrm{DMD}}(\mathbf{x}_t,t)
    =
    \mathbf{v}_{\mathrm{real}}(\mathbf{x}_t,t)
    -\mathbf{v}_{\mathrm{fake}}(\mathbf{x}_t,t).
\end{equation}
The student is updated using the detached pseudo-target
\begin{equation}
    \mathcal{L}_{\mathrm{DMD}}
    =
    \left\|
    \mathbf{G}_{\theta}
    -\operatorname{sg}\!\left(
    \mathbf{G}_{\theta}-\mathbf{g}_{\mathrm{DMD}}
    \right)
    \right\|_2^2.
\end{equation}
The fake-score model is separately trained with a flow-matching objective on
student-generated samples. We perform one student update and five fake-score
updates every six optimization iterations.

The distribution-matching stage additionally uses latent-anchor,
boundary-consistency, perceptual, and robot-focused auxiliary objectives:
\begin{equation}
\begin{aligned}
    \mathcal{L}_{\mathrm{SF\text{-}DMD}}
    ={}&
    \mathcal{L}_{\mathrm{DMD}}
    +\lambda_{\mathrm{anchor}}\mathcal{L}_{\mathrm{anchor}}
    +\lambda_{\mathrm{boundary}}\mathcal{L}_{\mathrm{boundary}} \\
    &+\lambda_{\mathrm{LPIPS}}\mathcal{L}_{\mathrm{LPIPS}}
    +\lambda_{\mathrm{robot}}\mathcal{L}_{\mathrm{robot}}
    +\lambda_{\Delta\mathrm{robot}}
      \mathcal{L}_{\Delta\mathrm{robot}},
\end{aligned}
\end{equation}
where the corresponding weights are 0.005, 0.05, 0.005, 0.005, and 0.1.

The robot-focused objectives use binary masks constructed by projecting robot
links, gripper branches, and end-effectors into each camera view using forward
kinematics and camera calibration. The masks are temporally expanded over
neighboring frames and spatially dilated at the latent resolution. Robot-region
reconstruction and temporal-motion losses are normalized by the valid mask
area. The left- and right-wrist losses are assigned twice the weight of the
head-view loss.

\subsubsection{Block-Autoregressive Rollout}

After each block is generated, its clean key-value features are inserted into
the causal cache. The next block is initialized from independent Gaussian noise,
while temporal continuity is maintained through the cached history rather than
by repeating the final frame of the previous block. For long-horizon rollout,
the cache retains one persistent sink block and the three most recent generated
blocks. This fixed four-block history bounds memory and computation as the
generated sequence becomes longer. The three camera views are generated jointly
within each block.

\subsection{Inference Configuration}
\label{app:inference_configuration}

Stage-II WorldLine uses 35 denoising steps, whereas Causal WorldLine uses four
steps for each generated block. Both models use an explicit Euler solver for
the rectified-flow probability-flow ODE and do not use classifier-free
guidance. Unless otherwise specified, inference uses a random seed of 42, a
per-view resolution of $480\times640$, and a batch size of one trajectory. The
head, left-wrist, and right-wrist views are generated jointly rather than being
concatenated into a single image input.

A 129-frame rollout consists of one observed initial frame followed by eight
generated blocks. Latency is measured end-to-end on a single NVIDIA H200 GPU
using FP16 precision and batch size one. The measurement includes VAE encoding,
video generation, and VAE decoding, while excluding MP4 encoding and disk
writing. Stage-II WorldLine requires 90 seconds for a 129-frame rollout,
corresponding to 0.698 seconds per frame. Causal WorldLine requires 21 seconds,
corresponding to 0.163 seconds per frame and a $4.3\times$ speedup.

\section{Experimental Details}

\subsection{Action-Conditioned Video Generation}
\label{app:action_generation_evaluation}

\noindent\textbf{Evaluation sets.}
We evaluate action-conditioned video generation on the scene-disjoint
AgiBotWorld test set and the fully out-of-domain DROID test set. The
AgiBotWorld test scenes and trajectories are excluded from training. DROID is
not used for WorldLine training, adaptation, or checkpoint selection and
differs from the predominantly bimanual training data in both camera viewpoints
and robot embodiment. Each method receives the same initial observation and
action sequence and predicts the corresponding future observations. CTRL-World
and Masked Visual Actions are evaluated using models trained on DROID.

\noindent\textbf{Full-frame metrics.}
PSNR, SSIM, and LPIPS are computed frame by frame using their standard
implementations, including the conditioning frame at $t=0$. Each metric is
averaged temporally within each trajectory and view, then equally across
trajectory--view pairs from the head, left-wrist, and right-wrist cameras.

\noindent\textbf{Robot-mask IoU.}
Robot-mask IoU is evaluated only on the head-camera view. Unlike wrist cameras,
which move together with the end effectors and often contain partial or
close-range views of the robot, the head camera provides a stable global view
of the workspace and more consistently captures the spatial relationship
between the robot and manipulated objects. It therefore enables a more
comparable measurement of action-conditioned robot motion across trajectories.

We use SAM~3 to segment the robot in each generated and ground-truth video,
using the end-effector projections only to initialize segmentation and then
propagating the masks over time. The two videos are segmented independently;
ground-truth masks are never used to prompt predicted videos. Given the
ground-truth and predicted robot masks $M_t$ and $\widehat{M}_t$, frame-level
IoU is
\begin{equation}
    \operatorname{IoU}_t
    =
    \frac{|M_t\cap\widehat{M}_t|}
         {|M_t\cup\widehat{M}_t|}.
\end{equation}
Frames in which both masks are empty are excluded rather than assigned an IoU
of one. If only one mask is empty, the frame receives an IoU of zero. We first
average valid frame-level IoUs within each video and then average over the
evaluation videos. Videos whose masks are empty in every frame are excluded
from this average.

\subsection{Embodied Planning}
\label{app:embodied_planning}

\noindent\textbf{Policies and tasks.}
We evaluate embodied planning in RoboTwin using one fixed checkpoint of
$\pi_{0.5}$ and one fixed checkpoint of LingBot-VLA, each trained on RoboTwin.
In contrast, WorldLine is directly evaluated in RoboTwin without using any
RoboTwin data for training, adaptation, or checkpoint selection. The policy
checkpoints are held fixed across all simulators and rollout budgets. They are
not required to represent fully converged policy performance because the
purpose of this experiment is to measure the relative improvement obtained
from simulator-based trajectory selection over direct execution of the same
policy. We evaluate 10 manipulation tasks with five scenes per task.

\noindent\textbf{Candidate generation.}
For each task and scene, a policy generates a bank of up to 32 candidate action
trajectories. Every simulator receives the same initial observation and
candidate actions. WorldLine renders each candidate as a predicted video. We
sample 16 frames from each video: ten frames are distributed over the complete
trajectory, while six additional frames are distributed over its final 40\%.
This sampling preserves the overall action progression while providing denser
evidence for placement, release, object dropping, and terminal-state
evaluation.

\noindent\textbf{Hierarchical candidate selection.}
We use Qwen3VL-8B to select one trajectory from each candidate bank. Candidate
identities are hidden from the model and replaced with temporary letter labels.
The candidate ordering is deterministically shuffled for each scene. For the
32-candidate setting, candidates are processed using a hierarchical bracket:
Qwen3VL-8B selects two candidates from each group of four, progressively
reducing the bank from 32 to 16, from 16 to 8, and from 8 to 4. The four
finalists are then scored independently from 0 to 100 according to their
predicted likelihood of completing the task. The highest-scoring candidate is
selected. Tied finalists are compared again using anonymous labels.

For each of five runs, each policy independently samples a new bank of 32
candidate trajectories using a distinct generation seed. All simulators
evaluate the same bank within each run. Rollout budgets of 2, 4, 8, 16, 24,
and 32 use prefixes of the run-specific candidate ordering and the corresponding
selection procedure. Reported error bars denote standard deviation across the
five independently sampled banks. Ground-truth success labels are accessed only
after selection and are never provided to Qwen3VL-8B.

\noindent\textbf{Qwen3VL-8B configuration.}
Qwen3VL-8B is evaluated in BF16 using deterministic decoding with
\texttt{do\_sample=False} and at most 256 generated tokens. The multimodal
conversation contains a single user message and no separate system prompt. The
following prompt is used for each four-way comparison. The value of
\texttt{select\_count} is two during the bracket rounds and one during a
tie-breaking comparison.

\begin{promptbox}{Prompt P1: Candidate Comparison}
Task instruction: {instruction}
Candidate A: <video A>
Candidate B: <video B>
Candidate C: <video C>
Candidate D: <video D>

Compare these candidates and select the {select_count} most likely to actually complete
the task. Prioritize correct final object state, then valid grasp/contact and placement.
Reject partial progress, drops, and physical inconsistencies. Ignore visual quality unless
it prevents judging. You must return exactly {select_count} candidate label(s), even if
every candidate appears to fail; in that case choose the least unsuccessful. Reply only
as JSON using the displayed letter labels:
{"selected_labels": ["A"], "reason": "<brief comparative reason>"}
\end{promptbox}

The final four candidates are scored independently using the following prompt.
The integer score is used as the primary ranking criterion, while
\texttt{task\_completed} and the textual fields are retained for analysis.

\begin{promptbox}{Prompt P2: Independent Candidate Scoring}
Task instruction: {instruction}
Candidate video:

Judge this candidate independently. Determine whether the predicted trajectory ends
with the task actually completed. Check the initial and final object state, valid
grasp/contact, correct placement, drops, and physical inconsistencies. Mere contact or
partial progress is not completion. Ignore visual quality unless it prevents judging.
Give a 0-100 task-success score. Reply only as JSON:
{"task_completed": <true or false>, "score": <integer>,
 "final_state": "<brief description>", "failure_reason": "<brief reason or empty>"}
\end{promptbox}

\noindent\textbf{Selector validation on ground-truth rollouts.}
To isolate ranking from video-generation errors, we apply Qwen3VL-8B directly
to ground-truth RoboTwin rollouts from $\pi_{0.5}$ using the same selection
protocol. Selected-trajectory success reaches 69.0\%, 79.3\%, 82.8\%, 82.8\%,
85.9\%, and 89.3\% for $N\in\{2,4,8,16,24,32\}$, respectively. This increase
shows that the selector can identify successful trajectories when given
accurate visual outcomes, supporting its use in rollout-based planning.

\subsection{Policy Evaluation}
\label{app:policy_evaluation}

\noindent\textbf{Evaluation set.}
Policy evaluation uses balanced sets of successful and failed robot
trajectories. For RoboTwin, we sample 50 base trajectories, comprising 25
successful and 25 failed executions drawn from $\pi_{0.5}$ and LingBot-VLA.
For AgiBot, we likewise sample 50 trajectories with an equal number of
successful and failed executions. All visual simulators are evaluated on the
same matched base trajectories. RoboTwin data are not used for WorldLine
training, adaptation, or checkpoint selection.

For each base trajectory, every simulator generates five outcome videos using
distinct generation seeds. Qwen3VL-8B decoding and the subsequent decision
rules remain deterministic, so variation across runs arises from video
generation.

WorldLine receives the initial observations and recorded action sequence and
generates the corresponding future video. WorldLine itself does not output a
success label. Instead, Qwen3VL-8B extracts structured visual evidence from
each generated trajectory, after which deterministic rules map the extracted
evidence to a success or failure prediction.

\noindent\textbf{Observation sampling.}
For each trajectory, we sample 12 temporal observations at normalized
positions
\begin{equation}
    p_i = \sqrt{\frac{i}{N-1}},
    \qquad i=0,\ldots,N-1.
\end{equation}
This schedule allocates more observations to the latter part of the trajectory,
where task completion and failure are typically more distinguishable. For
multi-view trajectories, all views are sampled at the same normalized
positions. The sampled images are resized to a height of 360 pixels. For grasp
tasks, synchronized views from the same time step are concatenated
horizontally. For lift and state-change tasks, synchronized views are provided
as separate image inputs in temporal order.

\noindent\textbf{Evidence-based trajectory evaluation.}
Instructions are routed into three task families: grasp, lift, and state
change. Qwen3VL-8B is instructed to extract observable evidence rather than
directly predict success or failure. In particular, wrist-camera motion alone
is not treated as evidence of object displacement; object motion must be
verified in the head view or through clear geometric separation from its
support.

The extracted evidence is converted into a binary prediction using
deterministic task-specific rules. A grasp requires the target to remain held
across at least two late observations. A lift additionally requires visible
target displacement in the head or wide view and separation from the original
support. A state-change task requires the instructed final state to be visibly
completed. A wrong object or visible drop results in failure, and insufficient
or null evidence is not interpreted optimistically.

Qwen3VL-8B uses BF16, deterministic decoding, at most 256 generated tokens, and
\texttt{enable\_thinking=False}. The following common prompt is used to extract
visual evidence. The output schema is selected according to the routed task
family.

\begin{promptbox}{Prompt P3: Visual Evidence Extraction}
Task instruction: {instruction}

Chronological observations follow. At each time step, the labeled views are
synchronized images of the same instant.

Extract observable facts only; do not decide success or failure. Track the target object
across the synchronized views and compare its initial, intermediate, and final states.
The hand_left and hand_right views are cameras mounted on the moving grippers: apparent
target motion, growth, or centering in a hand-camera view may be caused only by camera
motion and is not proof that the object moved. Verify object displacement in the wider
head view or by clear separation from its support and surrounding landmarks. Require
agreement between views when the head view is ambiguous.

Treat wording such as 'attempt to grasp X' as success only if X is actually grasped. For
a grasp-only task, require the target to remain visibly enclosed by the gripper across at
least two late observations; the target need not leave its support. For a lift or pick-up
task, require visible controlled motion of the object with the gripper and separation from
its support. Contact, gripper closure, or temporary occlusion alone is not sufficient.
State the exact time-step transition that proves any claimed motion. For placement,
insertion, folding, pouring, pressing, or turning, inspect whether the corresponding final
state is visible. Use null for a fact that is genuinely hidden in every view. Do not infer
facts from purposeful motion.

For press or click tasks:
A press or click is a momentary action: set required_final_state_visible true when the
instructed arm or gripper visibly reaches and presses the correct button, bell top, or
switch, even if it springs back or the surrounding object never moves. Mere approach
without visible contact or pressing is insufficient.

Return only the schema corresponding to the task family.

// grasp
{"target_held_for_two_late_steps": true,
 "wrong_object_or_drop": false,
 "observations": "...",
 "confidence": 0}

// lift
{"head_view_confirms_target_displacement": true,
 "target_held_for_two_late_steps": true,
 "target_on_original_support_at_end": false,
 "wrong_object_or_drop": false,
 "observations": "...",
 "confidence": 0}

// state_change
{"required_final_state_visible": true,
 "wrong_object_or_drop": false,
 "observations": "...",
 "confidence": 0}

Task instruction: {instruction}
Do not include a verdict.
\end{promptbox}

\noindent\textbf{Verification of unsuccessful predictions.}
When the initial evidence does not establish task completion, we apply a
conditional verification procedure. For synchronized multi-view trajectories,
we first inspect six higher-resolution observation groups at normalized
positions $[0,0.65,0.78,0.88,0.95,1]$, with images resized to a height of 480
pixels. This verification reuses the same evidence rubric and task-specific
schema.

For applicable state-change cases, we then check whether the already extracted
observations logically establish task completion without introducing new
visual claims.

\begin{promptbox}{Prompt P4: Logical Evidence Reconciliation}
Check the logical consistency of already-extracted visual observations; do not infer
anything that they do not explicitly state.
Task instruction: {instruction}
selected: {observations}
first pass: {observations}
high-resolution verification: {observations}
Do these observations themselves directly establish that the instructed task was
completed? A visibly completed momentary action such as pressing a button counts.
Placement, insertion, scooping, pouring, or transfer requires an explicitly observed
completed effect. Approach, contact, purposeful motion, an incomplete attempt, or a
negated completion does not count. Reply only with
{"observations_prove_completion": <true|false>, "reason": "<brief reason>"}
\end{promptbox}

Finally, unresolved lift and state-change cases are re-examined using the
temporal motion in the complete video. Twelve frames are sampled from each
available view and provided as temporally ordered video inputs.

\begin{promptbox}{Prompt P5: Temporal Video Verification}
Re-check the unsuccessful image-based judgment using the temporal motion in each camera
sequence. The views cover the same trajectory. Track the exact instructed target, support,
and destination. For lifting, require controlled target motion with the gripper and visible
separation or changed height relative to the support. For placement or insertion, require
the target to be released in the instructed destination; approach or temporary occlusion
is not completion.
Earlier observation: {observations}

Reply only with observable motion evidence:
{"task_effect_completed": <true|false|null>,
 "observations": "<exact visible transition; at most 30 words>"}
\end{promptbox}

A trajectory is counted as correctly evaluated when the final prediction agrees
with the recorded success or failure label. For each generation seed, accuracy
is computed by pooling the 100 trajectories from RoboTwin and AgiBot; the
reported 74\% is the mean across five seeds.

\subsection{Ablation Protocol}
\label{app:ablation_protocol}

All ablation variants use the same training-data splits, inference
configurations, evaluation cases, metric implementations, candidate
trajectories, and Qwen3VL-8B procedures as the corresponding main experiments.
For the planning ablations, each variant evaluates the same 32 candidate
trajectories generated by the fixed $\pi_{0.5}$ and LingBot-VLA checkpoints,
matching the $N=32$ setting in the main planning experiment. The reported
planning success rates are averaged over five evaluation runs.

For the no-action-free-data ablation, we exclude the Stage-I action-free corpus
entirely while matching the full WorldLine model's total training-compute
budget. Consequently, its comparison with the full model controls for total
compute and isolates the contribution of access to the action-free data more
directly than an unmatched-training baseline.

\section{Additional Quantitative Results}

\subsection{Representative-Task Planning Results}

Table~\ref{tab:appendix_representative_planning} reports planning performance
on a representative subset of RoboTwin tasks. The tasks are selected to cover
different manipulation types and interaction patterns, while excluding tasks
for which direct execution is saturated at either 0\% or 100\% for either
policy. The subset is not selected according to the magnitude of performance
improvement. All methods use the same policy checkpoints, candidate
trajectories, rollout budgets, and Qwen3VL-8B selector described in the
experimental protocol.

\begin{table*}[t]
    \centering
    \caption{Planning success rates on a representative subset of RoboTwin
    tasks (\%). The selected tasks cover different manipulation types. All methods evaluate the same candidate trajectories.}
    \label{tab:appendix_representative_planning}
    \scriptsize
    \setlength{\tabcolsep}{4.0pt}
    \resizebox{\textwidth}{!}{%
    \begin{tabular}{@{}lcccccc@{}}
        \toprule
        \textbf{Method}
        & \textbf{Handover Block}
        & \textbf{Lift Pot}
        & \textbf{Open Laptop}
        & \textbf{Stack Two Blocks}
        & \textbf{Turn Switch}
        & \textbf{Average} \\
        \midrule
        \rowcolor{LightOrange}
        \multicolumn{7}{c}{\textbf{$\pi_{0.5}$}} \\
        Direct Execution & 7.5 & 83.1 & 85.6 & 37.5 & 52.5 & 53.3 \\
        GE-Sim-V2 & 32.0 & 96.0 & 84.0 & 56.0 & 48.0 & 63.2 \\
        OpenDW-0.5 & 32.0 & 84.0 & 80.0 & 48.0 & 52.0 & 59.2 \\
        Masked Visual Actions & 0.0 & 100.0 & 80.0 & 68.0 & 56.0 & 60.8 \\
        \textbf{WorldLine} & 56.0 & 80.0 & 100.0 & 92.0 & 60.0 & 77.6 \\
        \textbf{WorldLine (Causal 4-Step)} & 8.0 & 84.0 & 100.0 & 64.0 & 72.0 & 65.6 \\
        \midrule
        \rowcolor{LightOrange}
        \multicolumn{7}{c}{\textbf{LingBot-VLA}} \\
        Direct Execution & 10.0 & 4.4 & 57.5 & 3.8 & 56.3 & 26.4 \\
        GE-Sim-V2 & 8.0 & 0.0 & 76.0 & 0.0 & 44.0 & 25.6 \\
        OpenDW-0.5 & 0.0 & 0.0 & 32.0 & 0.0 & 88.0 & 24.0 \\
        Masked Visual Actions & 8.0 & 0.0 & 20.0 & 4.0 & 60.0 & 18.4 \\
        \textbf{WorldLine} & 40.0 & 4.0 & 100.0 & 20.0 & 76.0 & 48.0 \\
        \textbf{WorldLine (Causal 4-Step)} & 8.0 & 0.0 & 88.0 & 36.0 & 88.0 & 44.0 \\
        \bottomrule
    \end{tabular}%
    }
\end{table*}

\noindent\textbf{Analysis.}
The task-level results show that the aggregate planning gains are not driven by
a single interaction type. For $\pi_{0.5}$, WorldLine increases the average
success rate over the five selected tasks from 53.3\% under direct execution to
77.6\%, a gain of 24.3 percentage points. The largest improvements occur on
Handover Block and Stack Two Blocks, where success increases by 48.5 and 54.5
points, respectively. For LingBot-VLA, WorldLine improves the subset average
from 26.4\% to 48.0\% (+21.6 points), with particularly large gains on Open
Laptop (+42.5 points) and Handover Block (+30.0 points). These improvements
span object transfer, articulated-object manipulation, and multi-object
stacking, supporting the use of predicted outcomes for candidate selection
across distinct interaction patterns.

The causal four-step variant retains substantial average gains over direct
execution, reaching 65.6\% for $\pi_{0.5}$ and 44.0\% for LingBot-VLA. Its
benefits are strongest on tasks with visually identifiable state transitions,
including Open Laptop, Stack Two Blocks, and Turn Switch. Performance is not
uniform across all tasks: neither WorldLine variant improves Lift Pot for
LingBot-VLA, and the causal variant provides little improvement on Handover
Block for $\pi_{0.5}$. This variation indicates that planning quality depends
on whether the predicted rollout exposes sufficiently discriminative evidence
for ranking the candidate actions.

\section{Additional Qualitative Results}

\subsection{Additional Comparisons on AgiBotWorld and DROID}

\begin{figure}[H]
    \centering
    \IfFileExists{Figures/agibot_droid_comparison_extra5.pdf}{%
        \includegraphics[width=\textwidth,height=0.68\textheight,keepaspectratio]{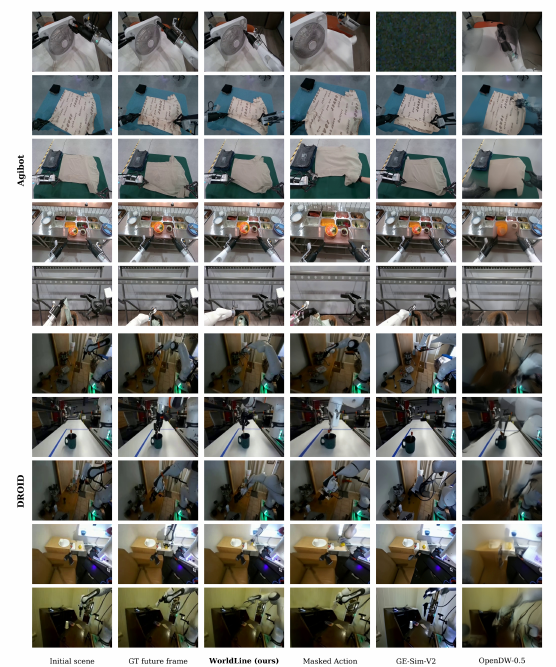}%
    }{%
        \fbox{\parbox[c][0.68\textheight][c]{0.96\textwidth}{
            \centering
            \textbf{Placeholder: Additional AgiBotWorld and DROID Comparisons}\\[1em]
            Initial observation $\mid$ action condition $\mid$ ground truth $\mid$
            baseline predictions $\mid$ WorldLine prediction\\[1em]
            Include enlarged views of action-critical robot--object interactions.
        }}%
    }
    \caption{\textbf{Additional qualitative comparisons on AgiBotWorld and
    DROID.} Each example shows the initial observation, action condition,
    ground-truth future, predictions from comparison methods, and the WorldLine
    prediction. AgiBotWorld examples are drawn from held-out scenes and include
    both successful and unsuccessful interactions, while DROID examples test
    transfer to unseen environments, viewpoints, and robot embodiments.
    Enlarged regions focus on commanded robot motion, contact, and the resulting
    object-state changes.}
    \label{fig:appendix_additional_comparisons}
\end{figure}

Figure~\ref{fig:appendix_additional_comparisons} provides additional
comparisons between WorldLine and prior video-generation and
action-conditioned simulation methods. We include held-out AgiBotWorld
examples covering both successful and unsuccessful interactions, together with
DROID examples involving unseen environments, camera configurations, and robot
embodiments. Each example uses the same initial observation and action sequence
for all methods. The visualization presents the ground-truth future alongside
the corresponding predictions and enlarges regions containing action-critical
robot--object interactions.

\subsection{Three-View WorldLine Predictions on AgiBotWorld}

\begin{figure}[H]
    \centering
    \IfFileExists{Figures/agibot_worldline_threeview_comparison.pdf}{%
        \includegraphics[width=\textwidth,height=0.9\textheight,keepaspectratio]{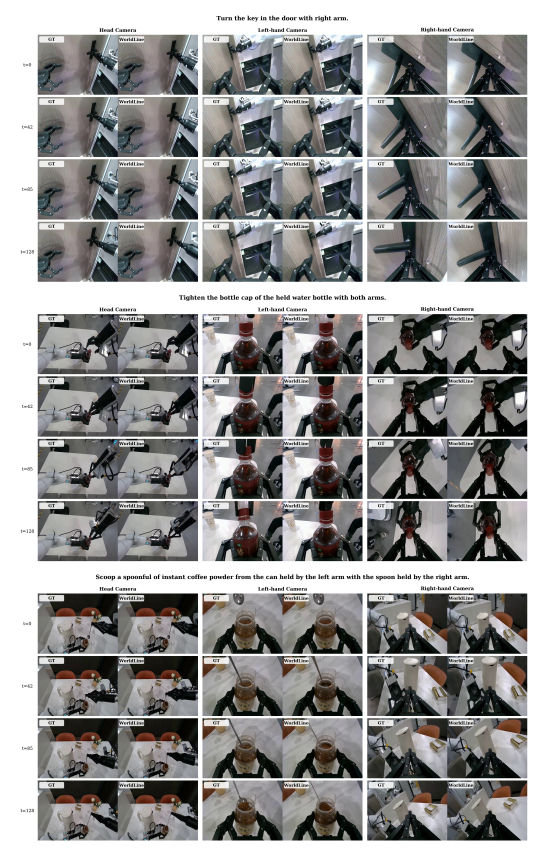}%
    }{%
        \fbox{\parbox[c][0.68\textheight][c]{0.96\textwidth}{
            \centering
            \textbf{Placeholder: AgiBotWorld Three-View WorldLine--Ground-Truth Comparisons}\\[1em]
            Temporally aligned head, left-wrist, and right-wrist views\\
            from ground-truth trajectories and WorldLine predictions.
        }}%
    }
    \caption{\textbf{Three-view comparison between WorldLine predictions and
    ground-truth AgiBotWorld trajectories.} Each example presents synchronized
    head, left-wrist, and right-wrist observations at multiple time steps from
    a held-out AgiBotWorld scene. The comparison shows robot motion, contact,
    and object-state evolution across the global and wrist-mounted camera
    views.}
    \label{fig:appendix_agibot_threeview}
\end{figure}

Figure~\ref{fig:appendix_agibot_threeview} provides additional comparisons
between WorldLine predictions and ground-truth trajectories from held-out
AgiBotWorld scenes. For each example, synchronized head, left-wrist, and
right-wrist views are shown at multiple time steps. This layout enables a
detailed examination of whether the predicted robot motion, contact, and
object-state changes remain temporally aligned across the global workspace
view and the two moving wrist cameras.

\subsection{Three-View WorldLine Predictions on DROID}

\begin{figure}[H]
    \centering
    \IfFileExists{Figures/droid_threeview_comparison.pdf}{%
        \includegraphics[width=\textwidth,height=0.86\textheight,keepaspectratio]{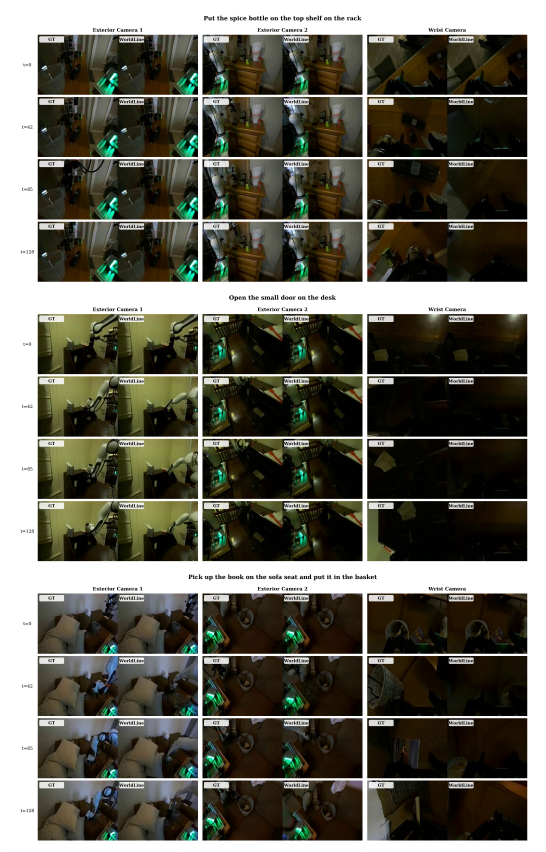}%
    }{%
        \fbox{\parbox[c][0.68\textheight][c]{0.96\textwidth}{
            \centering
            \textbf{Placeholder: DROID Three-View WorldLine--Ground-Truth Comparisons}\\[1em]
            Temporally aligned head, left-wrist, and right-wrist views\\
            from ground-truth trajectories and WorldLine predictions.
        }}%
    }
    \caption{\textbf{Three-view comparison between WorldLine predictions and
    ground-truth DROID trajectories.} Each example presents synchronized head,
    left-wrist, and right-wrist observations at multiple time steps. WorldLine
    is evaluated directly on DROID without training or adaptation on DROID
    data. The comparison exposes the temporal alignment of robot motion,
    contact, and object-state evolution across global and wrist-mounted camera
    views.}
    \label{fig:appendix_droid_threeview}
\end{figure}

Figure~\ref{fig:appendix_droid_threeview} presents additional comparisons
between WorldLine predictions and recorded DROID trajectories. DROID is not
used during WorldLine training, adaptation, or checkpoint selection. For each
example, we show temporally aligned head, left-wrist, and right-wrist views,
allowing the predicted robot motion and object evolution to be examined from
both the global workspace camera and the moving end-effector cameras. The
three-view layout also reveals whether the generated interaction remains
consistent across synchronized viewpoints under an unseen robot embodiment.

\subsection{Long-Horizon Causal Rollouts}

\begin{figure}[H]
    \centering
    \IfFileExists{Figures/worldline_causal_delayed_gt_threeview_comparison.pdf}{%
        \includegraphics[width=\textwidth,height=0.86\textheight,keepaspectratio]{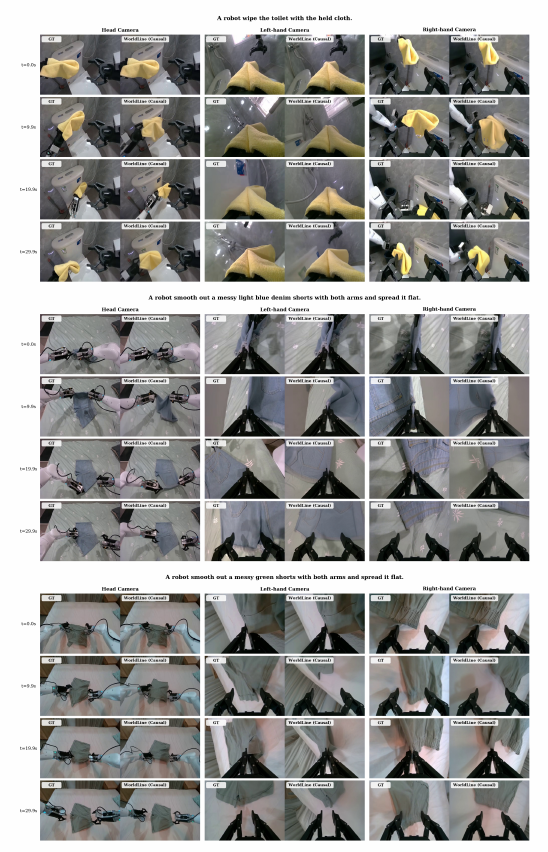}%
    }{%
        \fbox{\parbox[c][0.68\textheight][c]{0.96\textwidth}{
            \centering
            \textbf{Placeholder: Long-Horizon Three-View Causal Rollouts}\\[1em]
            Temporally aligned head, left-wrist, and right-wrist views\\
            from Causal WorldLine and the corresponding ground truth.
        }}%
    }
    \caption{\textbf{Long-horizon three-view rollouts generated by Causal
    WorldLine.} We compare the causal predictions with the corresponding
    ground-truth trajectories using synchronized head, left-wrist, and
    right-wrist observations sampled across the rollout. The visualization
    highlights long-term robot motion, object-state evolution, temporal
    continuity, and cross-view consistency under block-autoregressive
    generation.}
    \label{fig:appendix_causal_long_rollouts}
\end{figure}

Figure~\ref{fig:appendix_causal_long_rollouts} compares long-horizon Causal
WorldLine rollouts with the corresponding ground-truth trajectories. Each
example presents synchronized head, left-wrist, and right-wrist observations
sampled across the generated sequence. The visualization exposes how robot
motion, object-state changes, and cross-view consistency evolve across
successive block-autoregressive generation steps.

\subsection{Qualitative Ablations}

\begin{figure}[H]
    \centering
    \IfFileExists{Figures/agibot_worldline_ablation_comparison_compact_threecases.pdf}{%
        \includegraphics[width=\textwidth,height=0.75\textheight,keepaspectratio]{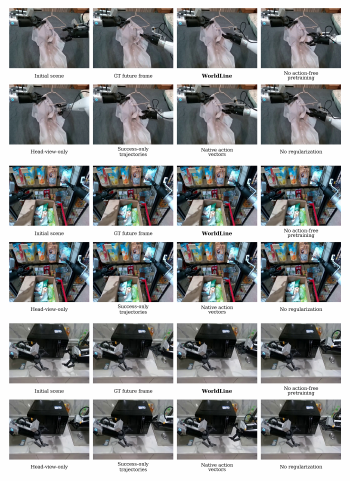}%
    }{%
        \fbox{\parbox[c][0.68\textheight][c]{0.96\textwidth}{
            \centering
            \textbf{Placeholder: Qualitative Ablation Comparisons}\\[1em]
            Initial observation $\mid$ action condition $\mid$ ground truth $\mid$
            full WorldLine $\mid$ selected ablation variants\\[1em]
            Organize examples by data, action control, and training strategy.
        }}%
    }
    \caption{\textbf{Qualitative effects of the main WorldLine components on
    AgiBotWorld.} Across three examples, we compare the full model with
    representative ablations of the training data, action-conditioning
    interface, and relational regularization. All variants receive the same
    initial observation and action sequence. Enlarged regions highlight
    differences in robot pose, contact, object motion, and the generation of
    unsuccessful interaction outcomes.}
    \label{fig:appendix_qualitative_ablations}
\end{figure}

Figure~\ref{fig:appendix_qualitative_ablations} visualizes the effects of the
main data, action-control, and training-strategy ablations on three AgiBotWorld
examples. Each comparison uses the same initial observation and action sequence
and places the ground-truth future, full WorldLine prediction, and selected
ablation variants side by side. The examples focus on differences in commanded
robot motion, contact accuracy, object displacement, and failure-state
generation that may not be fully captured by aggregate image metrics.

\clearpage

\clearpage

\clearpage

\section{Failure Cases}
\label{app:failure_cases}

\begin{figure}[t]
    \centering
    \includegraphics[width=\textwidth,height=0.72\textheight,keepaspectratio]{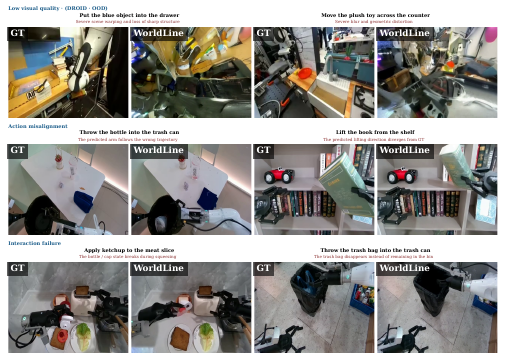}
    \vspace{-0.5cm}
    \caption{\textbf{Selected failure cases of WorldLine.} Each pair compares
    the ground-truth (GT) future with the WorldLine prediction. The top row
    shows two low-visual-quality examples from out-of-domain DROID evaluation;
    the middle row shows action misalignment; and the bottom row shows
    interaction failures involving inconsistent object or container states.}
    \label{fig:appendix_failure_taxonomy}
\end{figure}

Figure~\ref{fig:appendix_failure_taxonomy} shows selected WorldLine failures,
not estimates of how frequently each type occurs. The examples reveal residual
instability, which is especially apparent in the OOD DROID cases as severe
scene distortion and blur; other examples show action misalignment and
inconsistent interaction outcomes. This pattern motivates broader targeted
coverage of unseen environments and interactions, together with improved
robustness in action alignment and object-state prediction.

\section{Limitations and Broader Impact}

\subsection{Limitations}

Although WorldLine generalizes across unseen scenes, environments, and robot
embodiments, its coverage is still shaped by the diversity of interactions in
the training data. Rare manipulation patterns, deformable objects, unusual
camera configurations, and substantially longer action horizons may benefit
from broader and more targeted supervision. In downstream planning, the
achievable improvement is also bounded by the diversity of trajectories
proposed by the underlying policy and the quality of the multimodal selector.

\subsection{Broader Impact}

WorldLine can reduce the amount of repeated physical interaction required for
robot policy development by enabling candidate behaviors to be visualized,
evaluated, and selected before execution. This capability may lower data
collection costs, reduce hardware wear, support failure analysis, and make
cross-embodiment robot learning more accessible.

\end{document}